# A Compact and Efficient 1.251 Million Parameter Machine Learning CNN Model PD36-C for Plant Disease Detection: A Case Study

Shkëlqim Sherifi
*Computer Science*
*Independent Researcher*
Tetove, R.N.Macedonia
0009-0006-2227-5533

Shpend Ismaili
*Dept. of Computer Science*
*University of Tetovo*
Tetove, R.N.Macedonia
shpend.ismaili@unite.edu.mk

Florim Idrizi
*Dept. of Computer Science*
*University of Tetovo*
Tetove, R.N.Macedonia
0000-0001-7514-3282

Ejup Rustemi
*Dept. of Computer Science*
*University of Tetovo*
Tetove, R.N.Macedonia
ejup.rustemi@unite.edu.mk

Eip Rufati
*Dept. of Computer Science*
*University of Tetovo*
Tetove, R.N.Macedonia
eip.rufati@unite.edu.mk

***Abstract*— Deep learning has markedly advanced image-based plant disease diagnosis as improved hardware and dataset quality have enabled increasingly accurate neural network models. This paper presents PD36-C, a compact convolutional neural network (1,250,694 parameters and 4.77 MB) for plant disease classification. Trained with TensorFlow/Keras on the New Plant Diseases Dataset (87k images, 38 classes), PD36-C is designed for robustness and edge-deployability, complemented by a Qt-for-Python desktop application that offers an intuitive GUI and offline inference on commodity hardware. Across experiments, training accuracy reached 99.697% by epoch 30, and average test accuracy was 99.53% across 38 classes. Per-class performance is uniformly high, on lower end: Corn (maize)-Cercospora leaf spot with precision ~ 97,77% and recall ~ 0.9634, indicating occasional confusion with visually similar categories, while for the best cases on upper end: numerous classes including Apple Black rot, Cedar apple rust, Blueberry healthy, Cherry Powdery mildew, Cherry healthy, and all four grape categories achieves perfect precision 100% and recall of 1.00, indicating no false positive and strong coverage. These results show that with a well-curated dataset and careful architectural design, small CNNs can achieve competitive accuracy compared with recent baselines while remaining practical for edge scenarios. We also note typical constraints like adverse weather, low-quality imagery, and leaves exhibiting multiple concurrent diseases that can degrade performance and warrant future work on domain robustness. Overall, PD36-C and its application pipeline contribute a field-ready, efficient solution for AI-assisted plant disease detection in smart agriculture.***

## I. INTRODUCTION

Agriculture underpins human societies and economies, employing roughly one billion people (~28% of the global workforce), with the United States, China, and India among the leaders by net cropped area. Plant diseases account for an estimated 10-16% annual yield loss (~USD 220 billion) and contribute to persistent food contamination, making early detection and control a priority for global food security[1]. Cereal crops, wheat, rice, and maize constitute ~80% of total cereal production and supply about half of the world's calories. Rice is the primary staple across much of Asia, whereas maize plays a dominant role in the America [2].

Alongside parallel research in decentralized and secure data systems, our previous research demonstrated the effectiveness of CNN-based architectures in traffic monitoring contexts [3], [4]. Leveraging insights gained from these findings, we design a new CNN architecture tailored to plant disease detection in this research.

Across major cereals, a small set of pathogens drives a disproportionate share of losses. In wheat, rusts, septoria, powdery mildew, and Fusarium head blight are prevalent threats [5]. Rusts, fungal diseases manifesting as orange to red-brown pustules on leaves, can cause severe yield reductions, with wheat losses reported in the range of 45-50%, powdery mildew typically appears as a white to gray powdery growth on foliage [6]. In rice, common diseases include bacterial blight, leaf blast, brown spot, and tungro [7]. Maize (corn) is a critical crop for human consumption, livestock feed, and industrial use [8]. It is notably affected by leaf blight (a highly transmissible fungal disease), rust (often first visible as fine yellowish spots), gray leaf spot, and maize mildew, among other severe foliar diseases [9].

Timely detection, protection, and treatment are central to mitigating disease impacts and improving productivity. Biotic agents, including bacteria, fungi, and viruses, are primary drivers of low yields [10]. Compounding these technical challenges, structural shifts in rural labor and urban migration intensify on-farm constraints, reinforcing the need for scalable, field-ready diagnostic [10]. Traditional pest and disease diagnosis in crops has relied heavily on farmer experience and manual scouting. While foundational, these approaches are labor-intensive, prone to bias, slow to scale, and lack the predictive rigor required for consistent, efficient disease control [11]. Precision agriculture, by contrast, emphasizes early detection and prompt, accurate interpretation of plant health signals [12]. Although certain diseases may be latent or exhibit subtle early symptoms, many produce characteristic manifestations within the visible spectrum, particularly on leaves, creating an opportunity for image-based screening and decision support [10].

Laboratory diagnostics such as immunoassays, molecular-genetic identification, and related methods offer high specificity and sensitivity but are costly, resource-intensive, and often inaccessible to smallholders or routine field operations [13]. This motivates computational alternatives leveraging machine learning (ML) and computer vision (CV). Classical image analysis (e.g., thresholding, edge detection, K-means clustering, region-based segmentation) and ML classifiers (e.g., support vector machines - SVM) have been deployed alongside modern deep learning (DL) methods, particularly convolutional neural networks (CNNs), to quantify disease severity and enable automated recognition [14]. Given that most foliar diseases express visually on leaves, lightweight smartphone applications can, in principle, support in-person diagnosis and treatment recommendations at the point of need.

ML and, increasingly, DL offer speed, real-time responsiveness, and high predictive precision, automating repetitive detection, tracking, and classification tasks in crop protection [14], [15], [16]. However, state-of-the-art DL models typically demand large labeled datasets and substantial compute; their parameter counts inflate training time, and performance may degrade in small, heterogeneous, or in-the-wild datasets, raising concerns of overfitting and limited generalization [17]. Transfer learning (TL) provides a practical remedy, adapting pretrained models to agricultural imagery and better capturing spatial, spectral, and environmental dependencies under diverse weather and field conditions, often outperforming models trained from scratch [18]. Despite rapid progress, persistent gaps remain in robust feature extraction for fine-grained symptoms, resilience to environmental variability, and mitigation of overfitting, especially when moving from controlled datasets to operational field settings [19].

This study synthesizes current findings on plant disease detection with an emphasis on DL-based CV methods and their deployment characteristics, and proposes a compact, efficient, and high-precision CNN tailored for edge deployment, ultimately integrated into a ready-to-use Windows application.

The purpose of this article is to answer the following research questions:

**RQ1**: Why is plant disease detection of significant importance in the area of agriculture?

**RQ2**: To what extent can DL models address key challenges in plant disease detection, and what approaches yield efficient, compact ("***tiny***") models?

**RQ3**: Which model architectures are most suitable for accurate, robust foliar disease recognition?

**RQ4**: Can DL models operate reliably on resource-constrained edge devices in offline settings?

**RQ5**: How do ML/DL methods compare with traditional diagnostic practices in accuracy and reliability?

**RQ6**: Do current models produce stable predictions across diverse crops, diseases, and environments, and what are the main limitations?

The remainder of this paper is organized as follows: **Section 2** reviews related literature. **Section 3** details the methodology, including datasets, model architectures, and algorithms. **Section 4** reports model performance. **Section 5** describes the application design and edge-deployment considerations. **Section 6** discusses findings, limitations, and future directions. **Section 7** concludes with the paper's primary contributions.

## II. LITERATURE REVIEW

### A. Related Work

Deep learning (DL) has rapidly become the dominant paradigm for image-based plant disease diagnosis. Recent surveys highlight state-of-the-art convolutional neural networks (CNNs), transformer-based architectures, and generative adversarial networks (GANs), alongside modern imaging/analysis tools and evaluation protocols, as the core technical strands in the field [14]. Within this ecosystem, transfer learning (TL) with canonical CNN backbones (e.g., AlexNet, GoogLeNet, VGG, ResNet) has delivered consistently strong results across multiple crops and diseases.

***Transfer learning***. Early work using pretrained architectures reported high-accuracy classification on standard benchmarks: GoogLeNet, AlexNet, VGG, and AlexNet-OWTBn reportedly achieved 99.34% accuracy on leaf disease detection across fourteen plant species [20]. Combining AlexNet with GoogLeNet for four apple leaf diseases yielded 97.62% accuracy [21]. On tomato leaves, InceptionV3, VGG19, VGG16, and ResNet achieved 93.70% field accuracy [22].

***Architectures.*** To address complex field conditions, cluttered canopies, occlusions, and small, low-contrast lesions, attention augmented CNNs have been proposed. A lesion-focused self-attention CNN (SACNN) with a global-feature backbone attained 95.33% accuracy on AES-CD9214 and 98.0% on MK-D2, underscoring the value of attention mechanisms for field imagery [23].

***Systems and pipelines.*** Broader surveys have reviewed AI and IoT pipelines for tomato, chili, potato, and cucumber, covering prevalent diseases and symptoms, canonical processing flows, representative ML/DL models and datasets, and emerging directions such as edge-AI and drone-based, large-scale field monitoring [24]. Such systems complement model-centric studies by emphasizing deployment constraints and sensing modalities.

***Dataset-centric and augmentation.*** On the widely used PlantVillage dataset, a CNN trained on 20,636 images spanning tomatoes, pepper, and potatoes achieved 98.29% training accuracy and 98.029% test accuracy across 15 classes (12 diseased, 3 healthy) [25]. Motivated by the economic importance of tomato production in Mexico, another study combined public and in-field images with GAN-based augmentation to mitigate overfitting, reporting >99% accuracy on both training and test sets for tomato leaf disease classification [26]. Broader multi-species modeling across PlantVillage has also been reported: [27] modeled 12 crop species across multiple pathogen types (bacterial, viral, mold, mite) and healthy leaves, employing SVM/GLCM/CNN and K-means on real-time images, with accuracies of 98% (rice), 96% (apple), and 97% (tomato), evaluated via precision, recall, and F1. Methodologically, [28] proposed an enhanced CNN (ECNN+GA), showing advantages of deep networks and transfer learning with accuracies of 80% (TL), 85% (CNN+TL), 90% (ANN), and 95% (ECNN+GA), and [29] compared SVM, KNN, RF, LR, and CNN, finding RF best among classical ML at 97.12%, while CNN reached 98.43%.

***Optimization.*** Several recent works interrogate generalization under background/acquisition variability in rice leaf disease recognition. DenseNet169 achieved 99.66% test accuracy with TL, while fine-tuned Xception reached 99.99% [30]. A meta heuristic ACO-CNN integrating ant colony optimization with CNN features outperformed CGAN, plain CNN, and SGD, with 99.98% accuracy and the highest precision/recall/F1 (reported upper bounds ~99.6%, 99.97%, and 85% for the comparators) [31]. PPLCNet combined dilated convolutions, multi-level attention (CBAM), global average pooling (GAP), and weather-aware augmentation to reach 99.7% accuracy and 98.4% F1, supported by CAM-based interpretability [32]. A compact 2D CNN with two max-pool layers and fully connected heads attained 96% accuracy for tomato leaf diseases, surpassing SVM, VGG16, InceptionV3, and MobileNet baselines in that study [33].

***Comparative studies, real-time deployment, and mobile readiness.*** Extending beyond rice and tomato [34] developed an end-to-end pipeline for early ginger disease detection from 7,014 expert-labeled field images, achieving 95.2% test accuracy and delivering a deployable mobile application. Moving from classification to on-leaf detection and identification, TL-based object detectors are increasingly emphasized for field viability. For soybean insect detection, YOLOv5 achieved 98.75% accuracy at 53 FPS, outperforming InceptionV3 and a conventional CNN (97%), and contributed a multi-device, labeled dataset to facilitate robust deployment. [35]. Comparative evaluations of the superiority of DL over classical ML in several crops. For citrus leaf disease, DL outperformed ML, with VGG16 reaching 89.5% vs SVM at 87%, SGD at 86.5%, and RF at 76.8% [36]. Lightweight and embedded deployments have garnered attention: a hybrid wrapper feature selection with 2D-DWT features and a compact CNN was deployed on a Jetson Nano-equipped UAV for real-time, high-accuracy classification of apples, grapes, and tomatoes. [37]. On PlantVillage, fine-tuning multiple state-of-the-art CNNs showed DenseNet121 achieving 99.75% with fewer parameters. [38]. A 14-layer DCNN trained on an augmented set (BIM, DCGAN, NST) of 147,500 images attained 99.9655% accuracy with ~99.8% weighted precision/recall/F1, outperforming TL baselines. [39]. TL from ImageNet with VGGNet and Inception yielded >91.83% validation accuracy and ~92% under complex backgrounds for rice. [40]. Training on segmented leaves improved generalization to independent tests (98.6% over 10 classes) and increased model confidence relative to full-image training. [41]. A survey of 100 CNN studies underscored DCNN effectiveness for early diagnosis while outlining open challenges and future directions. [42]. Deployable, compute-efficient families InceptionV3 (98.42%), InceptionResNetV2 (99.11%), MobileNetV2 (97.02%), and EfficientNet-B0 (99.56%) demonstrated strong accuracy with reduced computation and mobile compatibility on a 14-species, 38-class benchmark. [43] and scaling up to 87,848 images across 25 plants/58 classes yielded 99.53% best accuracy, supporting advisory and early-warning applications [44].

***Ensembles and fusion.*** To address overfitting and the limits of single-backbone feature extraction, ensemble strategies have been proposed. PDDNet-AE and PDDNet-LVE integrate multiple pretrained CNNs via transfer learning with advanced fusion, improving robustness over single-model baselines [45].

***Laboratory diagnostics.*** Traditional visual inspection and mycological analysis remain foundational but are slow and labor-intensive. Modern laboratory methods (immunodiagnostics, molecular-genetic assays, and mass spectrometry) offer rapid, sensitive alternatives and increasingly complement imaging-based AI systems in plant health pipelines [46].

### B. Research Gap and Motivation

Despite the remarkable results achieved by prior studies, where some studies focus on literature review, some on specific use cases and some on a large number of plant categories, several limitations persist in the current state-of-the-art plant disease CNN classification works:

- *Offline and Edge Deployment*: Existing research solutions mostly rely on cloud infrastructure or high-end hardware for training and testing, and the number of layers and parameters varies, but mostly use a high number, making them less suitable for deployment in resource-constrained environments.
- *User Interface and Usability*: Most of the other implementations lack intuitive user interfaces, which present a barrier for adoption by non-technical users.

## III. Methodology

This study addresses autonomous plant disease detection via ML, with an emphasis on leaf analysis and disease prediction. We adopt a mixed-methods approach that combines:

- *Qualitative synthesis* of the scientific literature to ground the problem, scope architectures, and surface deployment constraints, and
- *Quantitative experimentation* to develop, implement, and evaluate a compact neural network model using CNN under realistic conditions.

The research proceeds in two phases:

- *Exploratory Phase*: We conducted a comprehensive review of peer-reviewed journals, conference proceedings, and reputable digital libraries to establish the state-of-the-art and identify methodological gaps relevant to field deployment.
- *Experimental Phase*: We designed and implemented three model variants, including a compact, self-developed sequential CNN (~1.25M parameters) using TensorFlow with Keras. We then performed an empirical evaluation to quantify performance and assess feasibility for real-world applications.

To systematize the exploratory phase, we queried major scholarly databases and indexes, including Google Scholar, Scopus, MDPI, IEEE Xplore, and ScienceDirect, among others. The following queries were employed: *Domain & Task*: Plant AND (Monitoring OR Surveillance OR Management). *Methods*: (Artificial Intelligence OR Computer Vision OR Machine Learning OR Deep Learning). *Scope*: (Models OR Applications). *Context*: (Case Study OR Real-world Application OR Field Study). Searches were filtered by publication year where appropriate to prioritize recent advances. Titles, abstracts, and keywords were screened for relevance; full texts were reviewed for methodological approach, dataset characteristics, and evaluation methods.

This methodological framework ensures both theoretical grounding and empirical validation, supporting the development of an efficient, robust, and scalable AI-based plant disease detection solution.

## A. Dataset

TABLE I. DATASET TRAIN/VALIDATION

| Dataset Train/Validation | | |
|---|---|---|
| ***Statistic*** | ***Training Set*** | ***Validation Set*** |
| Unique_categories | 38 | 38 |
| Total Images | 70,295 | 17,572 |
| Average | 1,849.87 | 462.42 |
| Standard Deviation | 104.32 | 26.13 |
| Min | 1,642 | 410 |
| Max | 2,022 | 505 |
| Balance_indicator | 0.056 | 0.057 |

For training and evaluation, we employed the New Plant Diseases Dataset hosted on Kaggle [47]. This dataset is widely used in the CV community due to its diversity, scale, and consistent labeling, and it contains more images than PlantVillage (~54,305 images reported in [43], [41], [48]).

The dataset consists of approximately 87,900 RGB images of healthy and diseased crop leaves, with >65,000 labeled instances, spanning 38 classes and totaling roughly 1.43 GB of image data [8]. Each image is associated with an explicit class label, enabling supervised learning for plant disease recognition. Fig. 2 illustrates a representative sample from each class (left to right). To ensure consistency and compatibility with the proposed architecture, all inputs were standardized to a resolution of $256 \times 256$ pixels. The images cover a broad range of real-world conditions, including varied illumination, camera viewpoints, and challenging scenes with blur and background clutter.

Partitioning and class balance. The dataset was divided into 80% training (70,295 images), 20% validation (17,572 images), and a test set of 33 images, providing a balanced foundation for model development and final assessment.

The per-class averages are $1,850 \pm 104$ images for training and $462 \pm 26$ images for validation according to the standard deviation, Table I. As shown in Fig. 1, class distributions are well balanced, with a coefficient of variation of $\sim 6\%$, which lies comfortably within commonly recommended thresholds for split balance $< 30\%$.

This configuration provides a large, diverse, and well-balanced benchmark suitable for rigorous supervised training and comparative evaluation of plant disease detection models.

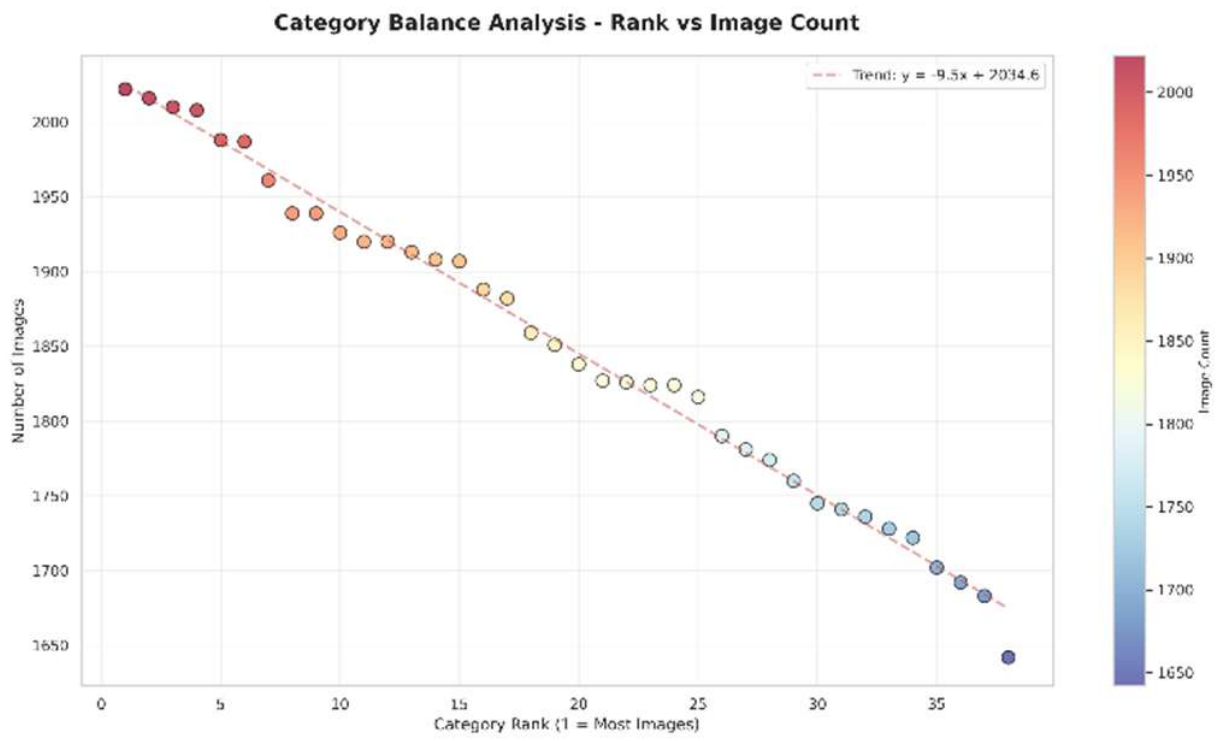

Fig. 1 Dataset categories spread number

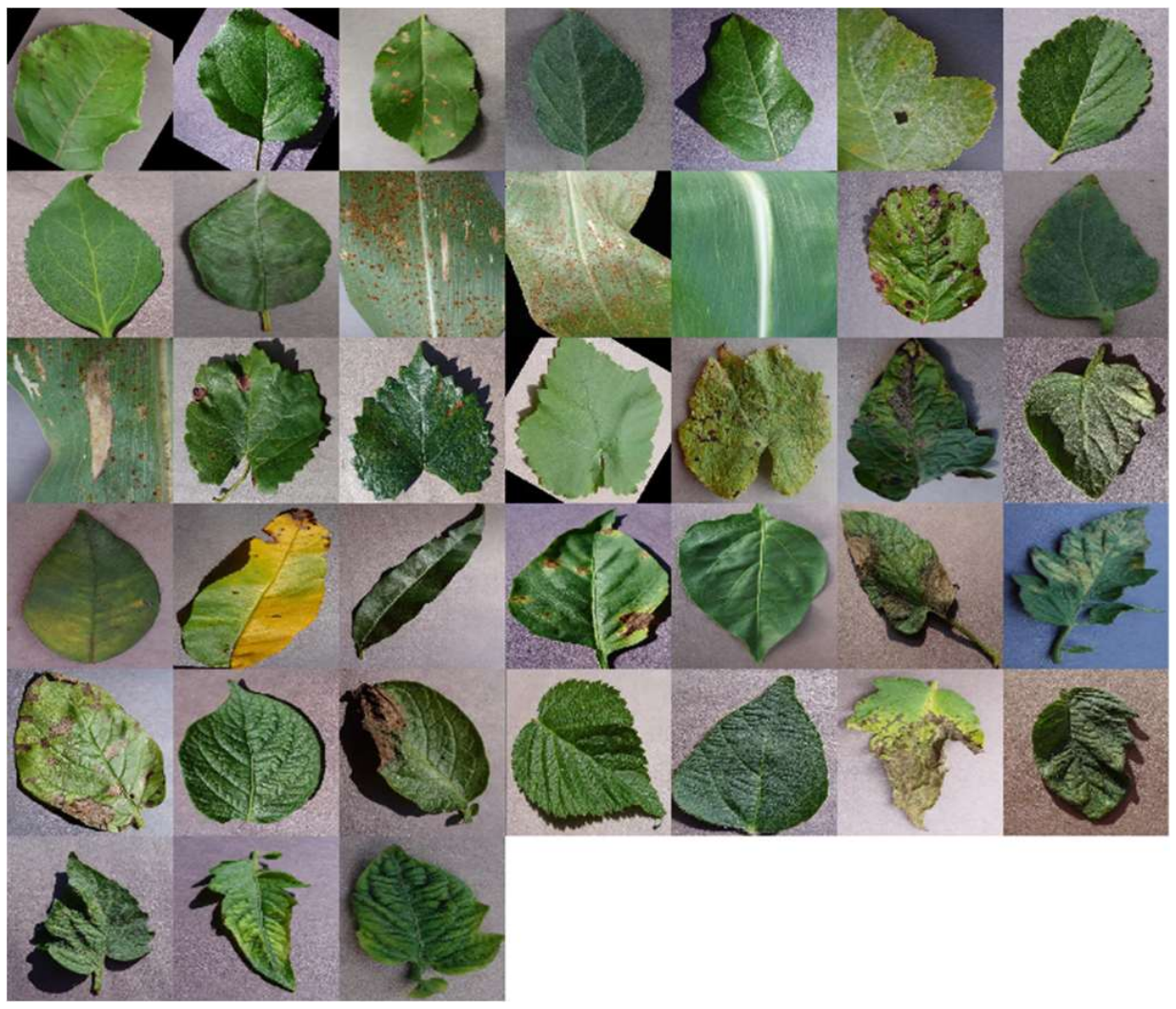

Fig. 2 Dataset sample images

## B. Data Augmentation

To improve generalization and reduce overfitting on the 38-class plant disease dataset, which, despite containing over 87,900 images, exhibits limited real-world acquisition diversity due to its controlled-environment origin, online data augmentation was applied stochastically to every training batch. All transformations were applied on a block named *plant_augmentation*, embedded as the first stage of the model graph.

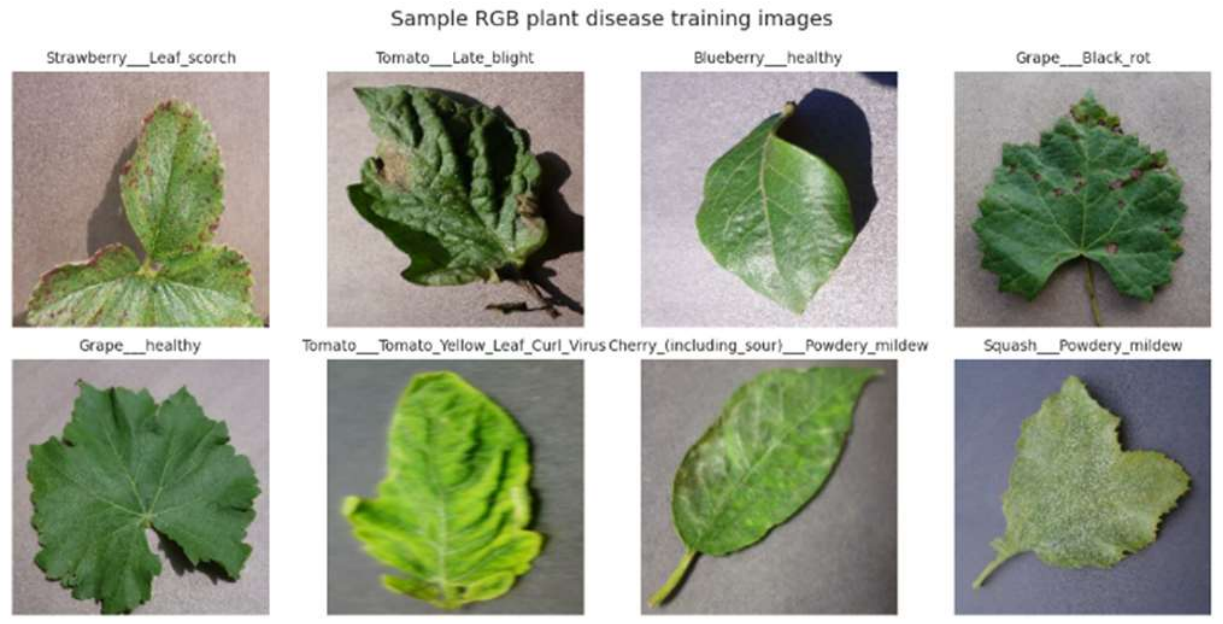

Fig. 3 Sample RGB plant disease training images

Fig.3 introduces representative samples from a single training batch of original RGB images at $224 \times 224$ resolution across eight disease classes, and Fig. 4 shows the same images after stochastic augmentation: subtle horizontal flips, rotations, translations, zoom, and contrast shifts while the morphology is preserved.

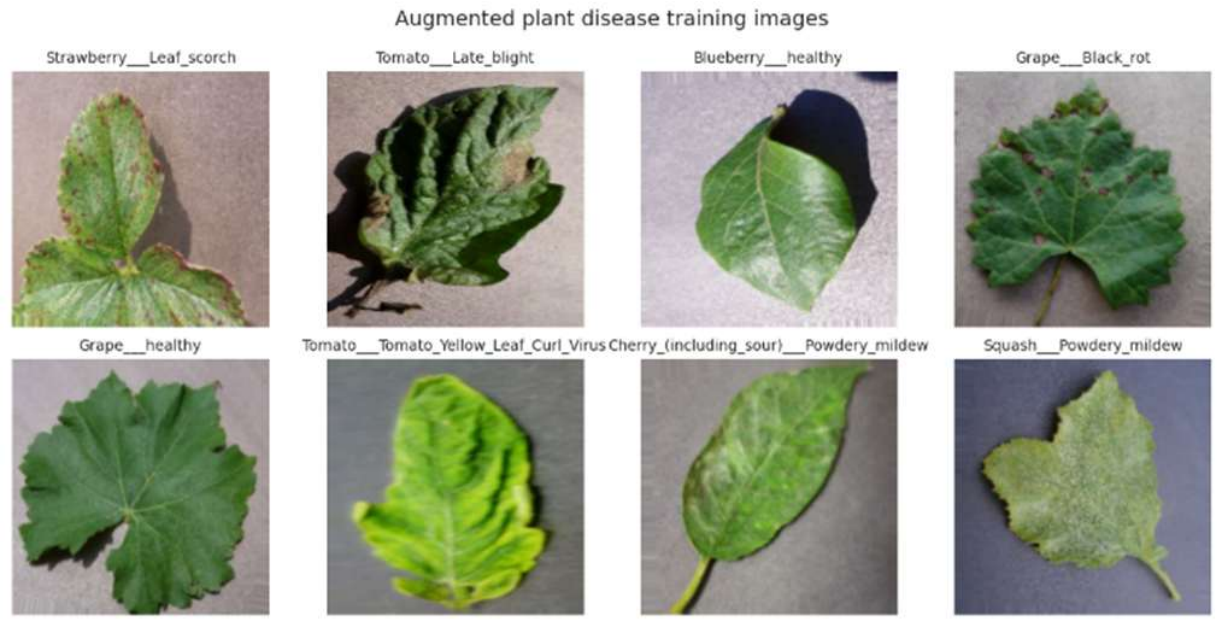

Fig. 4 Augmented plant disease training images

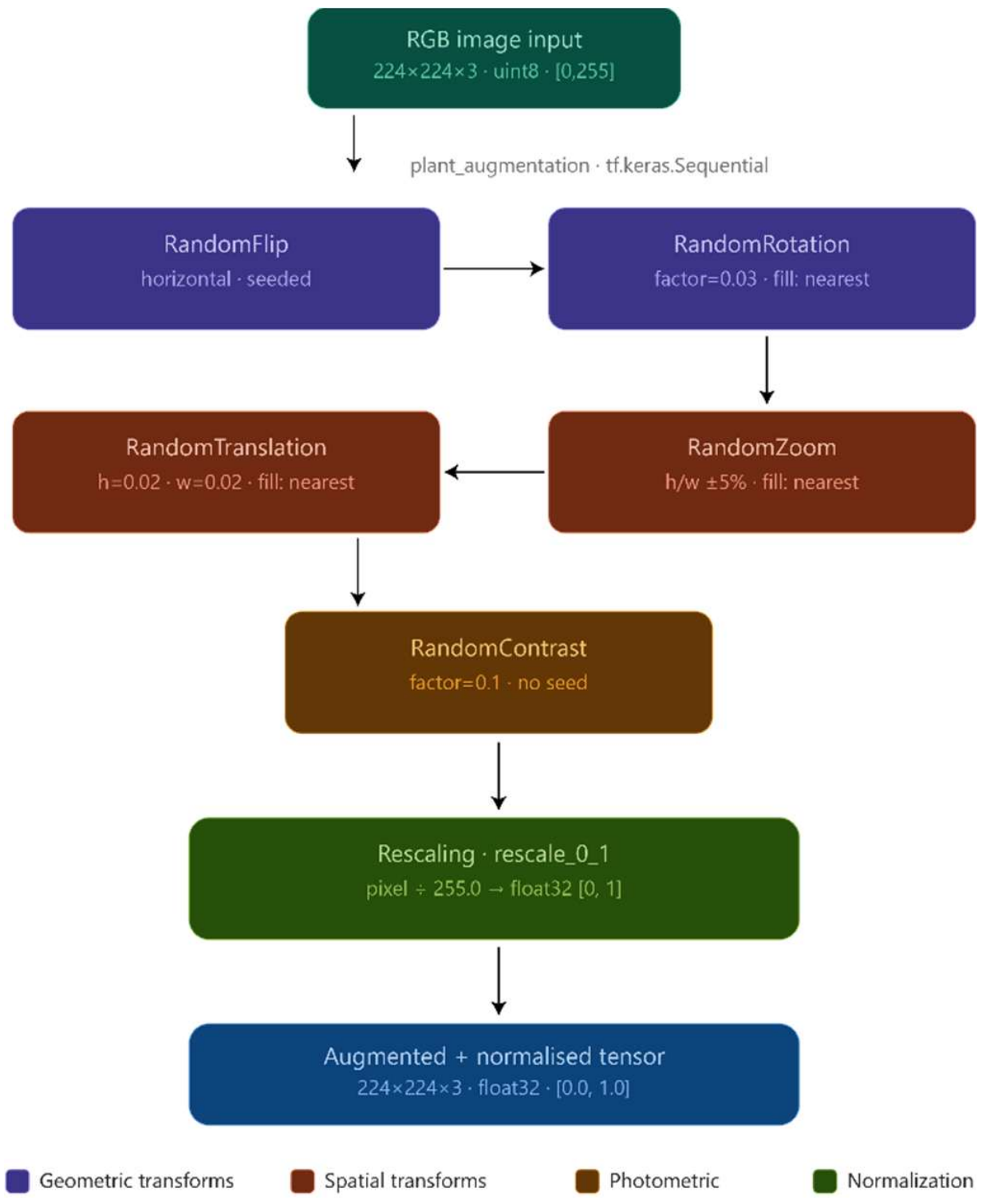

Fig. 5 Data augmentation pipeline

The pipeline applies five successive perturbations Fig 5. First, each image is randomly flipped horizontally with probability $x = 0.5$; Vertical flipping was deliberately excluded because leaf orientation in field images is predominantly upright. Second, a small random rotation of up to $\pm 3\%$ (approximately $\pm 10.8°$) is applied, which is sufficient to account for a slight camera tilt without introducing unrealistic leaf geometries. Third, random translation shifts the image by up to $\pm 2\%$ of its height and width, simulating minor framing variation between acquisitions. Fourth, random zoom shifts the spatial scale by $\pm 5\%$ in both height and width, emulating variation in shooting distance. All geometric and spatial transforms use nearest-neighbor fill to avoid introducing artificial blurred or zero-padded boundary artefacts at image edges, which could otherwise be learned as spurious diagnostic features. Fifth, random contrast adjustment with a factor of $x = 0.1$ is applied as a photometric transform to desensitize the model to mild illumination and camera response stochasticity independent of the geometric chain. Immediately following the augmentation block, a *Rescale* layer divides all pixel values by 255.0, mapping the uint8 input data type range $[0,255]$ to float32 $[0.0, 1.0]$. This normalization is mathematically equivalent to min-max scaling under the assumption that the sensor's full dynamic range is $[0,255]$, and ensures that the weight initialization schema and optimizer learning rate operate in a numerically stable range. The augmentation block is active only during the forward pass during training. At validation and inference time, the stochastic augmentation layers are bypassed to ensure evaluation metrics are computed properly.

## C. Architecture

A typical image-based classification pipeline includes image acquisition, preprocessing, region of interest (ROI) identification, feature extraction/selection, and final classification. Formally, the model maps a color image $xi \in R^{H \times W \times 3}$ to a probability vector $y_i \in R^C$ over $C$ predefined classes via a sequence of transformations (layers):

$$f_\theta : x_i \rightarrow y_i = \mathrm{softmax}(z_i), \quad z_i \in R^C \qquad (1)$$

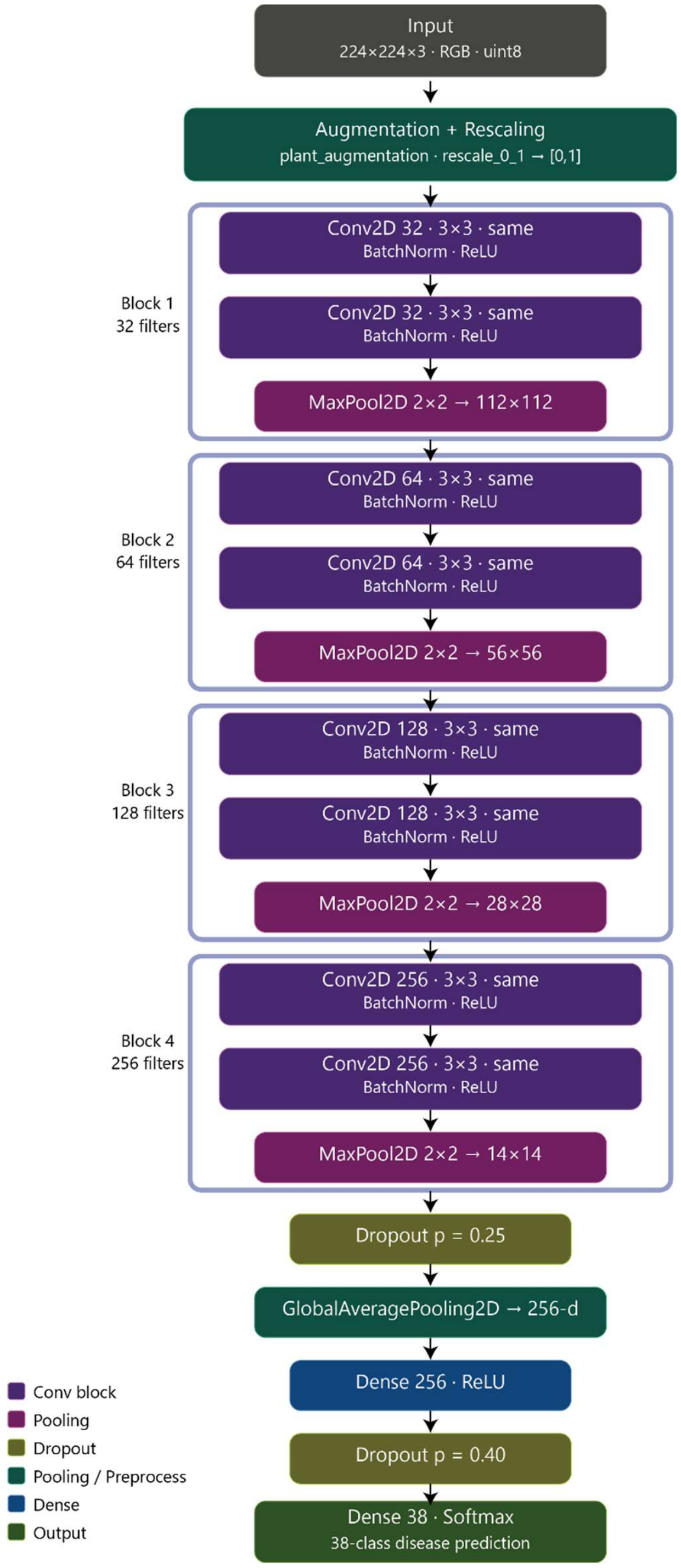

Fig. 6 PD36-C Architecture

The PD36-C network extends a conventional CNN by integrating augmentation, scaling, convolutional, pooling, dropout, flattening, and dense layers to improve both efficiency and accuracy. It is an 18-layer deep CNN with the following learnable/parametric layers (8 Conv2D layers, 8 BatchNorm layers, and 2 Dense layers) with dropout regularization, comprising *1,250,694* total parameters. Inputs are RGB images of size $224 \times 224$, for training, validation, and testing, and outputs are 38 logits corresponding to 38 plant-disease categories. Training uses the Adam optimizer with a learning rate of $1 \times 10^{-4}$ to $0.5 \times 10^{-5}$, and a batch size of 8; Fig. 6 illustrates the architectural layout.

PD36-C is organized into four convolutional blocks, each containing $3 \times 3$ kernel size followed by a MaxPooling2D with $2 \times 2$ kernel with stride = 2. The architecture exhibits a progressive feature hierarchy with filter depth increasing geometrically: 32 → 64 → 128 → 256.

After every pair of convolutions, a MaxPooling2D layer reduces spatial dimensions by a factor of 2, yielding a progression from $224 \times 224$ to the final $14 \times 14$ feature maps. The **ReLU** activation function is used after each convolution. These four blocks realize 8 convolutional layers and 4 pooling layers, implementing a progressive feature hierarchy with increasing channel depth and decreasing spatial resolution.

To enhance generalization and normalization (to prevent overfitting), PD36-C employs two dropout stages: ***Dropout-1*** with a rate $r_1 = 0.25$, and ***Dropout-2*** with a rate $r_2 = 0.40$.

The classification head proceeds as:

- ***Dropout-1*** ($r_1 = 0.25$)
- ***Flatten*** the final $14 \times 14$ feature maps into a 1-D.
- ***Dense-1*** with 256 neurons and ReLU activation.
- ***Dropout-2*** ($r_2 = 0.40$).
- ***Output Dense*** with 38 neurons using Softmax activation.

$$P = \sum_{i=1}^{20} P_i \quad (2)$$

The total parameter count is P, where $P_i$ denotes parameters in the layer $i$ (convolutional kernels, biases, and dense weights). PD36-C has: **1,170,432 (~93.6%)** Convolutional, and **65,792 (~5.3%)** Dense parameters. Table II presents Layers with their type, output shape, and number of parameters.

TABLE II. MODEL ARCHITECTURE PD36-C

| **Layers/types/Shapes/Parameters** | | | |
|---|---|---|---|
| ***Layer*** | ***Type*** | ***Output Shape*** | ***Parameter*** |
| plant_augmentation | Sequential | (224, 224, 3) | 0 |
| rescale_0_1 | Rescaling | (224, 224, 3) | 0 |
| conv2d | Conv2D | (224, 224, 32) | 864 |
| Batch_normalization | BatchNormalization | (224, 224, 32) | 128 |
| activation | Activation | (224, 224, 32) | 0 |
| conv2d_1 | Conv2D | (224, 224, 32) | 9,216 |
| Batch_normalization_1 | BatchNormalization | (224, 224, 32) | 128 |
| Activation_1 | Activation | (224, 224, 32) | 0 |
| max_pooling2d | MaxPooling2D | (112, 112, 32) | 0 |
| conv2d_2 | Conv2D | (112, 112, 64) | 18,432 |
| Batch_normalization_2 | BatchNormalization | (112, 112, 64) | 256 |
| Activation_2 | Activation | (112, 112, 64) | 0 |
| conv2d_3 | Conv2D | (112, 112, 64) | 36,864 |
| Batch_normalization_3 | BatchNormalization | (112, 112, 64) | 256 |
| Activation_3 | Activation | (112, 112, 64) | 0 |
| max_pooling2d_1 | MaxPooling2D | (56, 56, 64) | 0 |
| conv2d_4 | Conv2D | (56, 56, 128) | 73,728 |
| Batch_normalization_4 | BatchNormalization | (56, 56, 128) | 512 |
| Activation_4 | Activation | (56, 56, 128) | 0 |
| conv2d_5 | Conv2D | (56, 56, 128) | 147,456 |
| Batch_normalization_5 | BatchNormalization | (56, 56, 128) | 512 |
| Activation_5 | Activation | (56, 56, 128) | 0 |
| max_pooling2d_2 | MaxPooling2D | (28, 28, 128) | 0 |
| conv2d_6 | Conv2D | (28, 28, 256) | 294,912 |
| Batch_normalization_6 | BatchNormalization | (28, 28, 256) | 1,024 |
| Activation_6 | Activation | (28, 28, 256) | 0 |
| conv2d_7 | Conv2D | (28, 28, 256) | 589,824 |
| Batch_normalization_7 | BatchNormalization | (28, 28, 256) | 1,024 |
| Activation_7 | Activation | (28, 28, 256) | 0 |
| max_pooling2d_3 | MaxPooling2D | (14, 14, 256) | 0 |
| dropout | Dropout | (14, 14, 256) | 0 |
| global_average_pooling2d | GlobalAveragePooling2D | 256 | 0 |
| dense | Dense | 256 | 65,792 |
| dropout_1 | Dropout | 256 | 0 |
| predictions | Dense | 38 | 9,766 |
| | | | |
| **Trainable params** | | | 1,248,774 |
| **Non-trainable params** | | | 1,920 |
| **TOTAL** | - | - | **1,250,694** |

The proposed CNN architecture comprises 18 trainable layers organized into a backbone for feature extraction and a classification head for prediction generation, as illustrated in Fig. 6.

*1)* ***BACKBONE FEATURE EXTRACTOR****. The backbone progressively extracts hierarchical features from the input image ($I \in R^{224 \times 224 \times 3}$) through five convolutional blocks:*

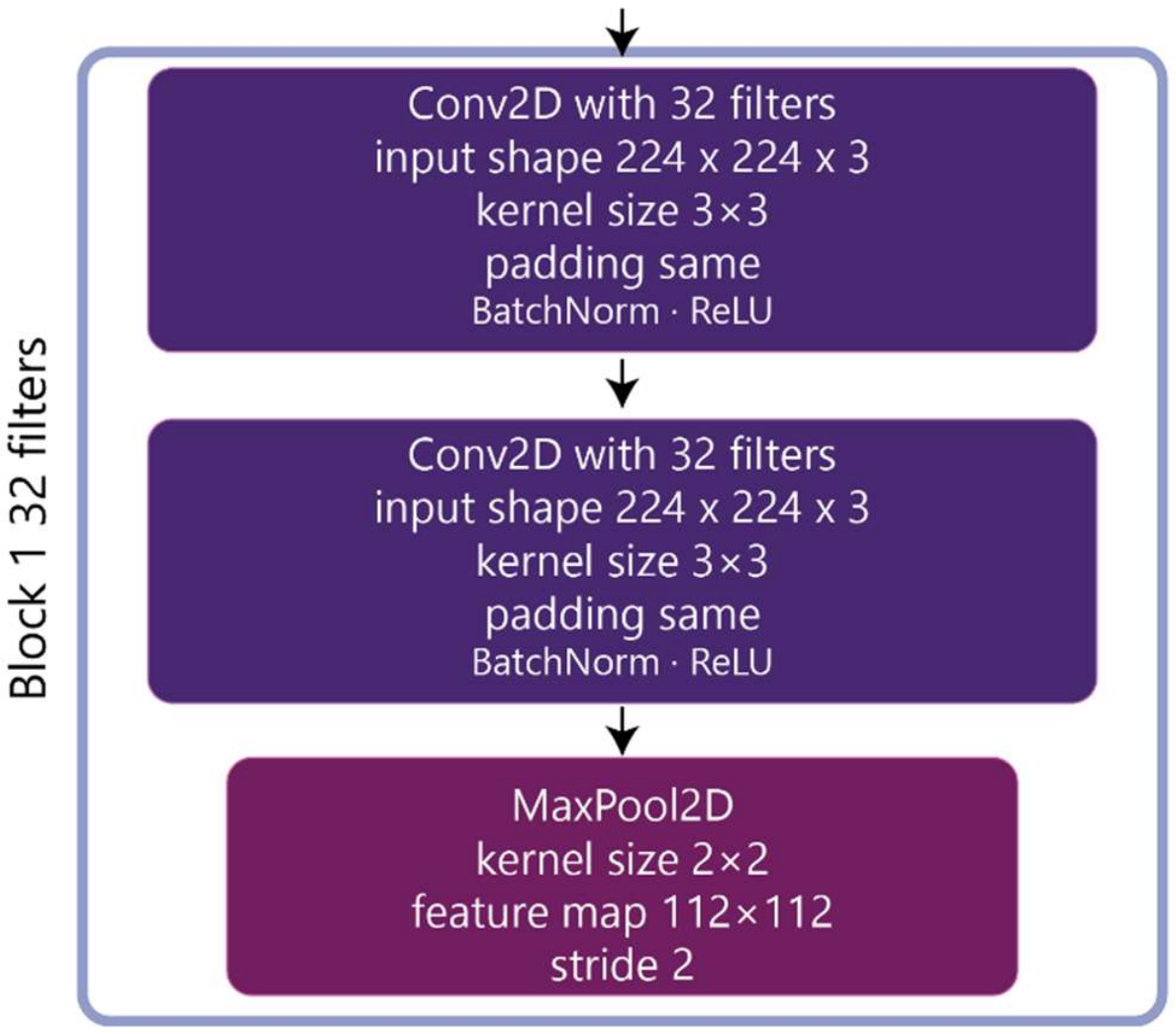

Fig. 7a PD36-C Block 1

- ***Block 1***

$$(X_1 = \text{Conv2D}(I, 32{,}3 \times 3{,}\text{padding} = \text{same,ReLU}))$$

$$(X_2 = \text{Conv2D}(X_1, 32{,}3 \times 3{,}\text{padding} = \text{same,ReLU})$$

$$(X_3 = \text{MaxPool2D}(X_2, 2 \times 2{,}\text{stride} = 2))$$

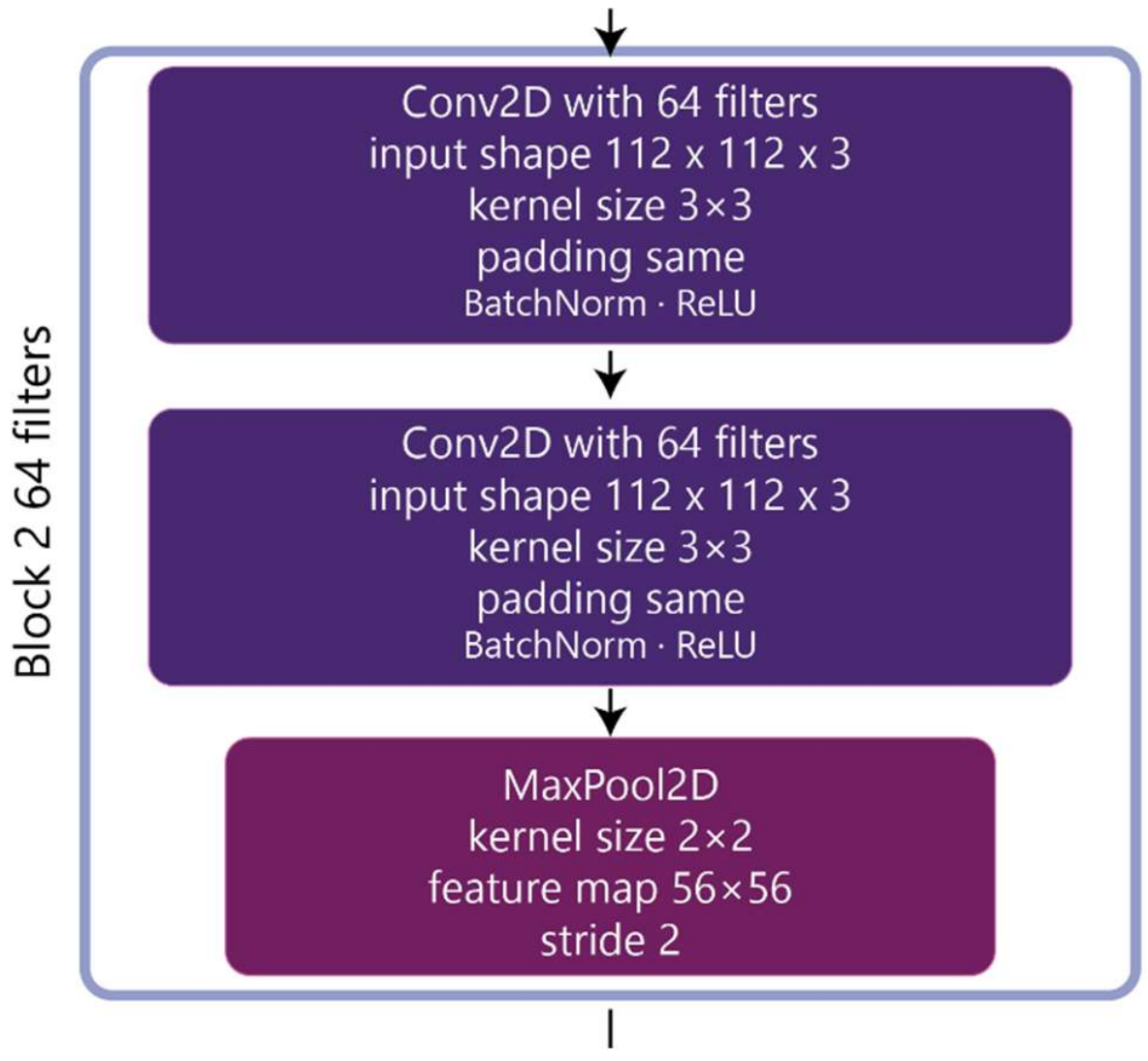

Fig. 7b PD36-C Block 2

- ***Block 2***

$$(X_4 = \text{Conv2D}(X_3, 64{,}3 \times 3{,}\text{padding} = \text{same,ReLU}))$$

$$(X_5 = \text{Conv2D}(X_4, 64{,}3 \times 3{,}\text{padding} = \text{same,ReLU}))$$

$$(X_6 = \text{MaxPool2D}(X_5, 2 \times 2{,}\text{stride} = 2))$$

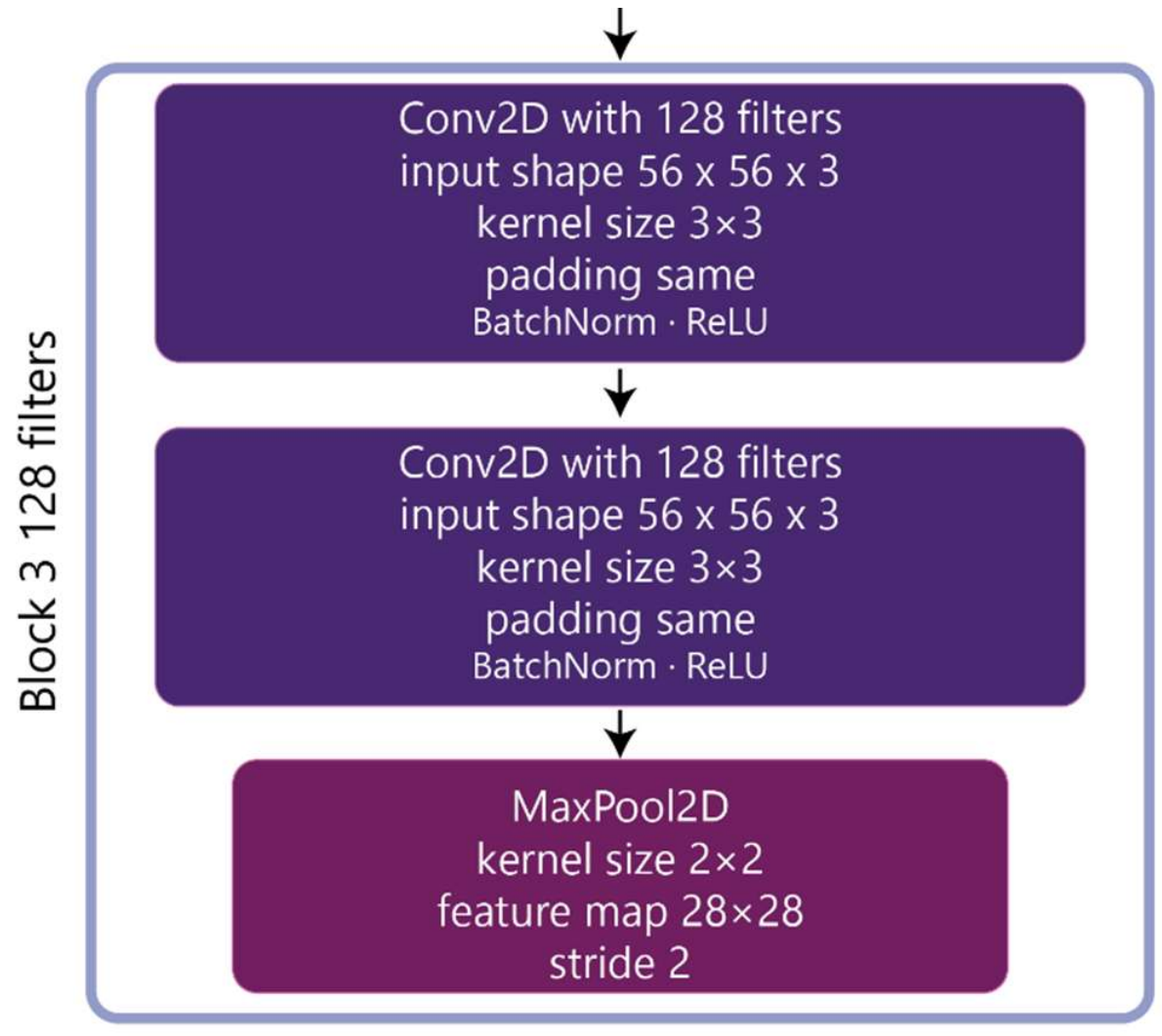

Fig. 7c PD36-C Block 3

- ***Block 3***

$$(X_7 = \text{Conv2D}(X_6, 128{,}3 \times 3{,}\text{padding} = \text{same,ReLU}))$$

$$(X_8 = \text{Conv2D}(X_7, 128{,}3 \times 3{,}\text{padding} = \text{same,ReLU}))$$

$$(X_9 = \text{MaxPool2D}(X_8, 2 \times 2{,}\text{stride} = 2))$$

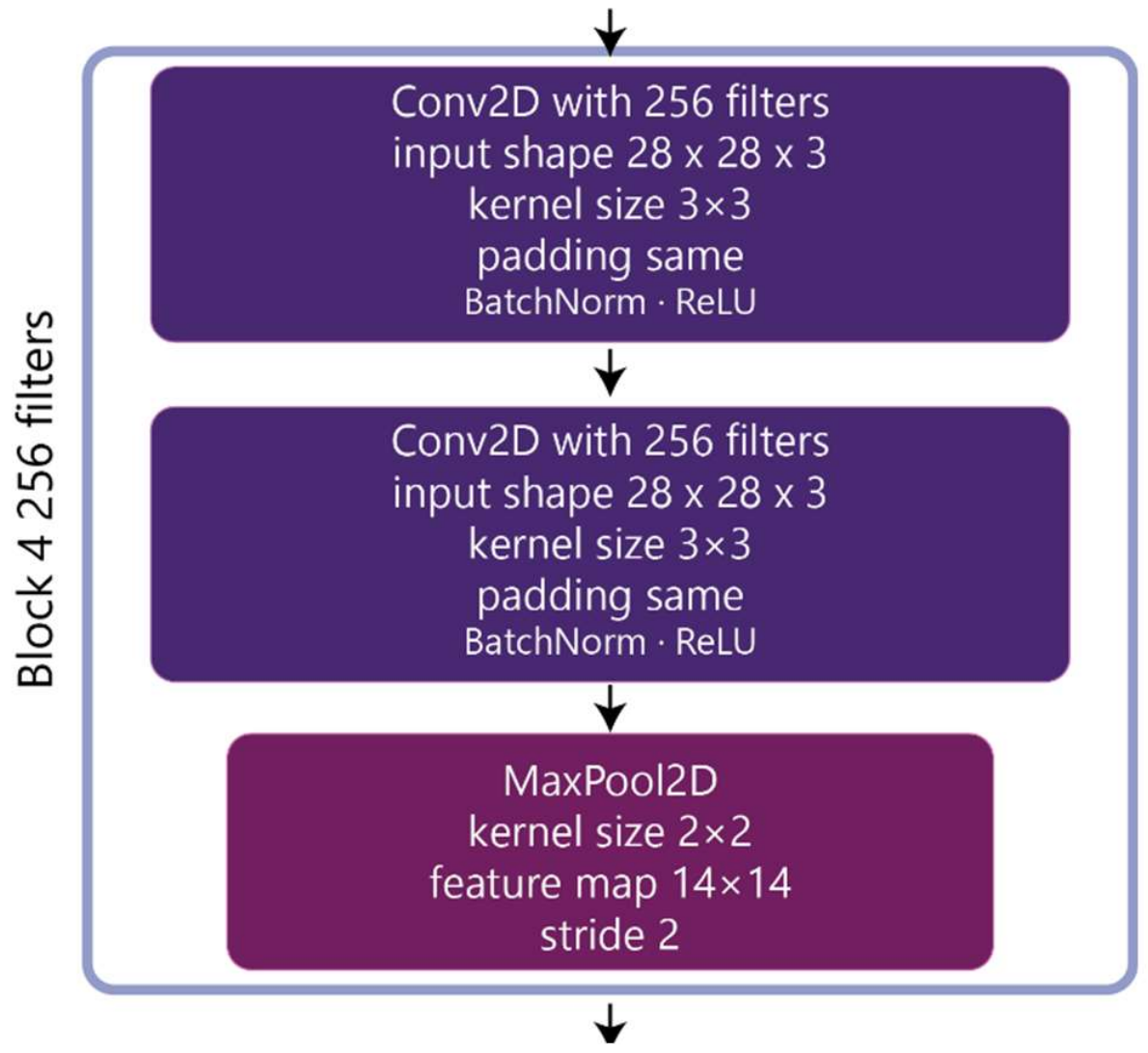

Fig. 7d PD36-C Block 4

- ***Block 4***

$$X_{10} = \text{Conv2D}(X_9, 256{,}3 \times 3{,}\text{padding same,ReLU}))$$

$$(X_{11} = \text{Conv2D}(X_{10}, 256{,}3 \times 3{,}\text{padding same,ReLU}))$$

$$(X_{12} = \text{MaxPool2D}(X_{11}, 2 \times 2{,}\text{stride} = 2))$$

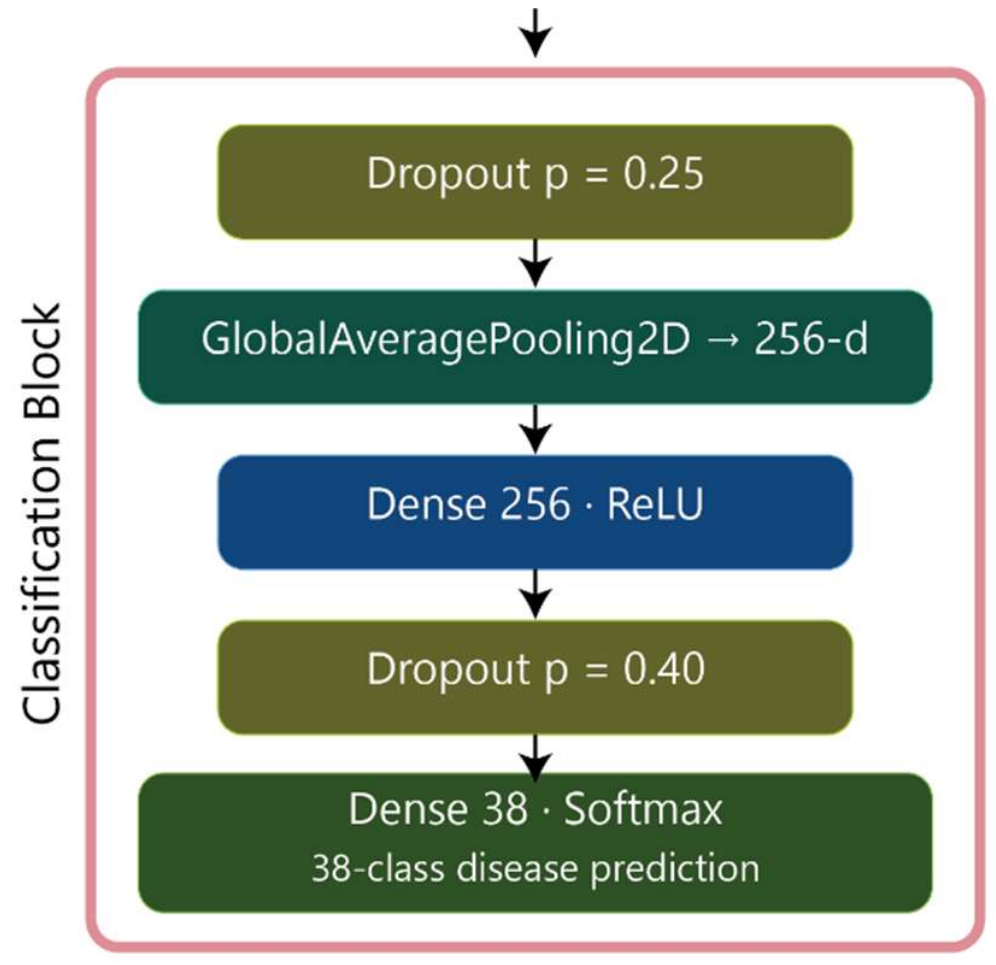

Fig. 8 PD36-C Classification Block

*2) **CLASSIFICATION HEAD**. The extracted features are transformed into class predictions through:*

- ***Regularization and Flattening***

$$(X_{13} = \text{Dropout}(X_{12}, r = 0.25))$$

$$(X_{14} = \text{Flatten}(X_{13}))$$

- ***Dense Layers***

$$(X_{15} = \text{Dense}(X_{14}, 256\text{,ReLU}))$$

$$(X_{16} = \text{Dropout}(X_{15}, r = 0.4))$$

$$(Y = \text{Dense}(X_{16}, 38\text{,softmax}))$$

where $(Y \in R^{38})$ represents the final class probability distribution.

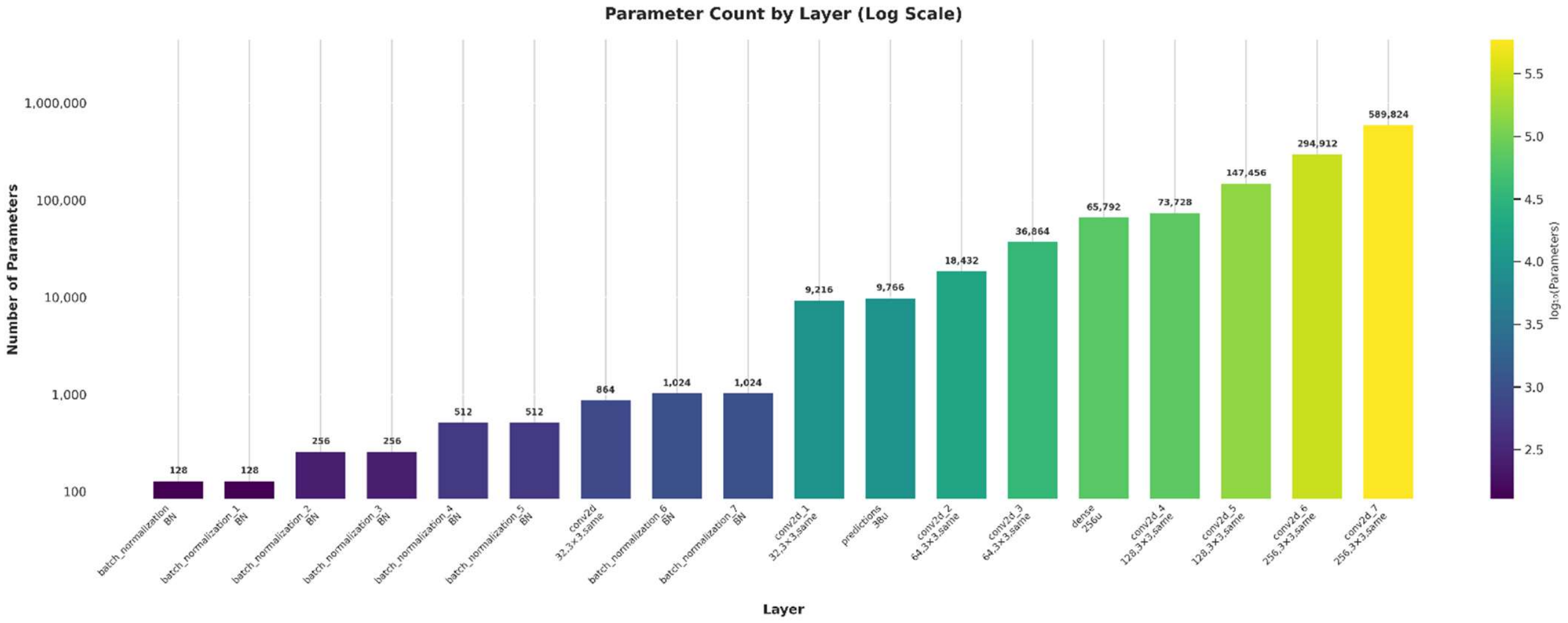

Fig. 9 PD36-C Number of parameters for each layer

Fig. 9 reports per-layer parameterization: Dense accounts for **65,792** parameters, while Conv6 and Conv7 contribute **294,912** and **589,824** parameters, respectively.

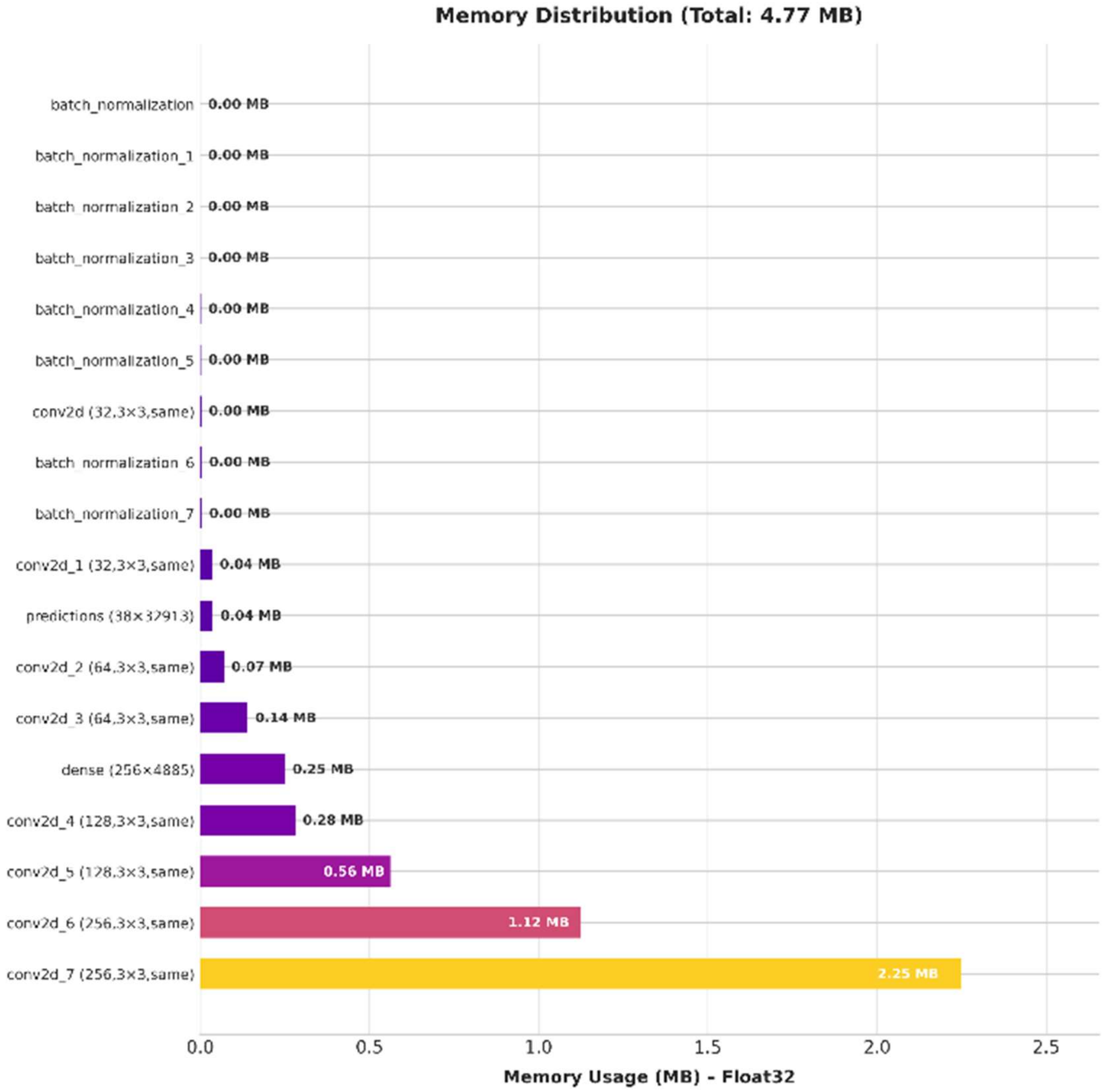

Fig. 10 PD36-C size for each layer

Fig. 10 presents layer-wise memory usage. Conv6 consumes approximately ~**1.12 MB**, followed by Conv7 at roughly **2.25 MB**. Table III summarizes a benchmark between PD36-A, PD36-B, and PD36-C.

TABLE III. MODEL ARCHITECTURE COMPARISON

| *Metrics* | *PD36-A [49]* | *PD36-B[49]* | *PD36-C[49]* |
|---|---|---|---|
| **Type** | Sequential CNN | Sequential CNN | Sequential CNN |
| **Characteristic** | Deep CNN | Deep CNN with Dropout Regularization | Deep CNN with Dropout Regularization |
| **Total layers** | 15 | 20 | 18 |
| **Total Parameters** | 17,989,446 | 7,917,894 | 1,250,694 |
| **Input shape** | 128x128x3 (RGB) | 128x128x3 (RGB) | 224x224x3 (RGB) |
| **Output shape** | 38 | 38 | 38 |
| **Optimizer** | Adam, LR=0.001 | Adam, LR=0.0001 | Adam, LR=0.00005-0.0001 |
| **Convolutional layers** | 8 | 10 | 8 |
| **Filter progression** | 32>64>128>256 | 32>64>128>256>512 | 32>64>128>256 |
| **Kernel size** | 3x3 | 3x3 | 3x3 |
| **Padding strategy** | 'same' | 'same' AND 'valid.' | 'same' |
| **Activation** | ReLU | ReLU | ReLU |
| **Pooling layers** | 4 | 5 | 4+1 |
| **Pooling type** | MaxPooling2D | MaxPooling2D | MaxPooling2D + GlobalAveragePooling2D |
| **Pooling size** | 2x2 | 2x2 | 2x2 |
| **Pooling strides** | 2 | 2 | 2 |
| **Dense layers(DL)** | 2 | 2 | 2 |
| **Hidden DL units** | 1024 | 1536 | 256 |
| **Output DL units** | 38 with Softmax | 38 with Softmax | 38 with Softmax |
| **Dropout layers(DOL)** | / | 2 | 2 |
| **DOL 1** | / | 25% | 25% |
| **DOL 2** | / | 40% | 40% |
| **Convolutional parameters** | 1,172,256(6.5%) | 4,712,224(59.5%) | 1,170,432 (~ 93.6%) |
| **Dense parameters** | 16,817,190(93.5%) | 3,147,264(39.7%) | 65,792 (~ 5.3%) |
| **Loss Function** | Categorical Cross-Entropy | Categorical Cross-Entropy | Categorical Cross-Entropy |
| **Traning Accuracy** | 0.9825 | 0.9818 | 0. 99697 |
| **Traning Loss** | 0.0555 | 0.0569 | 0. 44110 |
| **Forward Pass Operations** | ~ 35,978,892 FLOPs | ~ 15,835,788 FLOPs | / |
| **Memery** | ~ 68.62 MB | ~ 30.20 MB | ~ 14.2 MB |
| Project link: https://github.com/shkelqimsherifi/AI_DeepLearning_CNN_Model_Plant_Disease_Detaction_PUBLICversin.git | | | |

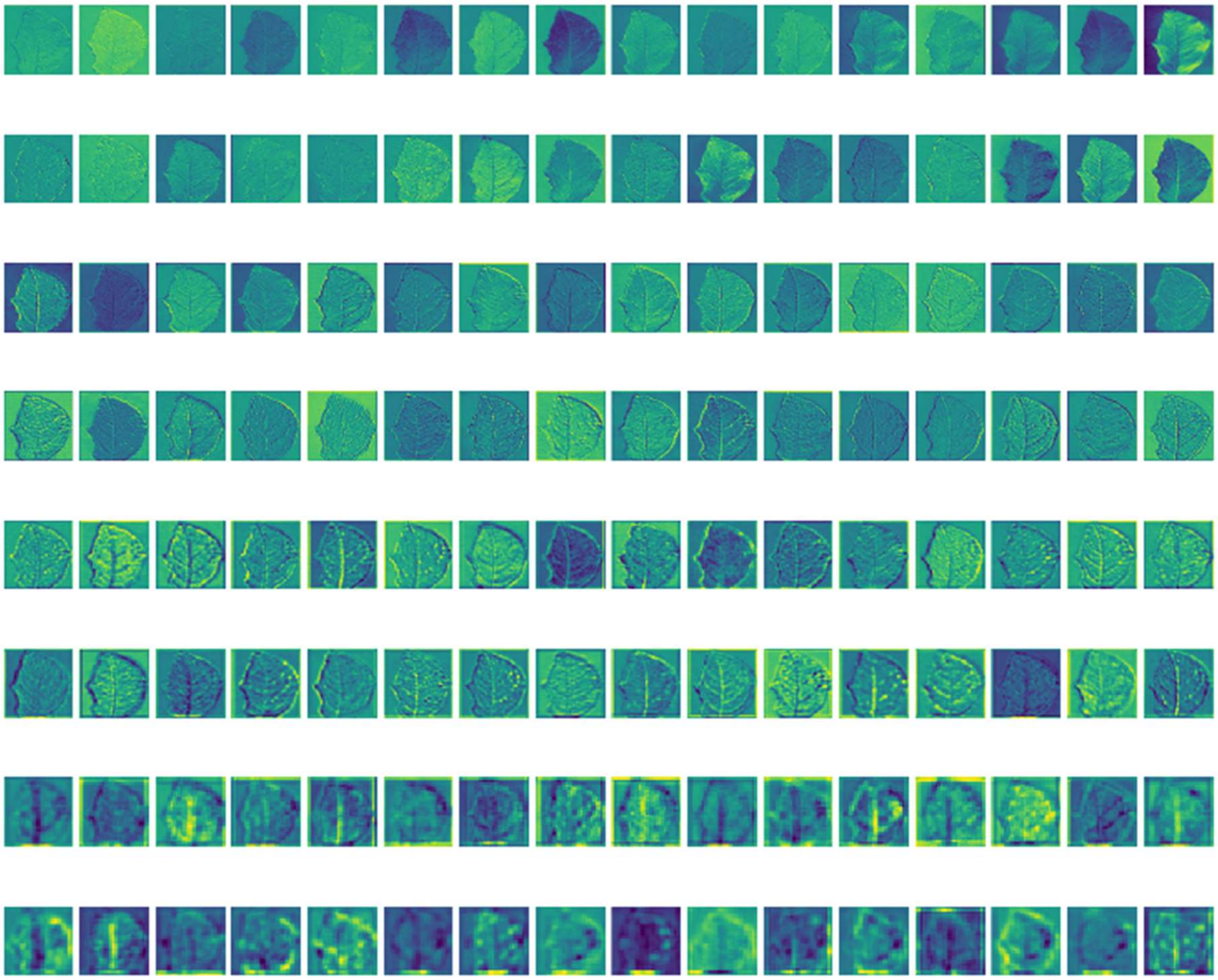

Fig. 11 Feature map activations extracted from each of the eight Conv2D layers of the trained plant disease CNN.

With an improved training strategy and deeper feature hierarchy, PD36-C:

- *Reduces parameters from ~18 M to ~1.252 M;*
- *Cuts memory from ~69 MB to ~4.77 MB;*
- *Improve accuracy and loss, indicating a more favorable accuracy-efficiency trade-off.*

The 18-layer PD36-C CNN implements a structured, progressively deep feature extractor with targeted dropout regularization and a compact classifier head, striking a balanced trade-off between representational capacity and computational efficiency for a medium-scale, 38-class plant disease classification model.

D. Model Feature Analysis

To examine how visual information is transformed across the hierarchy of the CNN, we construct a functional sub-model that simultaneously outputs the post-activation feature of all eight Conv2D layers. For the selected test image, each convolutional output tensor of shape H × W × F is cropped based on the filter dimension, and the first 16 filters are normalized. This visualization traces the progression from low-level edge and color detectors to disease-specific activations in the deepest layers. This provides qualitative evidence that the network extracts a meaningful hierarchical representation of plant pathology.

In Fig. 11, each row corresponds to one convolutional layer, and each column shows the response of one of the first 16 learning filters, normalized to [0,1] and rendered with a colormap. Early layers (rows 1-2) produce broad edge and color-gradient detectors, mid-level layers (rows 3-4) focus on texture and lesion-boundary detectors, and finally, deep layers (rows 5-8) produce highly sparse, semantically specific activations corresponding to disease regions.

E. Algorithms

The proposed model PD36-C employs supervised classification, a Multi-Layer Perceptron (MLP) for image-based plant disease recognition. The selected method was chosen after carefully trade-off analysis of probabilistic classification schemes of the following ML algorithms: Naive Bayes(NB), Decision Tree(DT), Nearest Neighbor(NN), Vector Machine Support(VMS), and Random Forest(RF) [48], [50].

The CNN model uses the following algorithms:

- *Convolution operation*:

$$(I * K)(i,j) = \sum_m \sum_n (i+m, j+n) K(m,n) \quad (3)$$

where $I$ is the input image, $K$ is the kernel, and (i, j) are the coordinates of the output.

- *ReLU activation*:

$$ReLU(x) = max(0,x) \quad (4)$$

- Pooling operation (Max pooling):

$$P(i,j) = \max_{m,n} I\,(i+m, j+n) \quad (5)$$

where $P$ is the pooled feature map.

- Feature extraction:

$$F = \text{PretrainedModel}(I) \quad (6)$$

where $F$ is the feature map extracted from the input $I$.

- Fully connected layer:

$$y = W \cdot F + b \quad (7)$$

IV. MODELS METRICS

The primary objective of our experiments is to evaluate the PD36-C plant disease detection model and analyze its effectiveness using standard quantitative metrics. We report accuracy, precision, and F1-score as primary metrics. Where applicable, we distinguish training/validation from test performance. Following model development, we conducted a series of tests whose outcomes are summarized in Tables V-VII. All images in the validation set were standardized to 256 × 256 pixels, for inference, inputs were resized to the model's native 224 × 224 × 3 format to maintain architectural compatibility. As expected for deep models, training is compute-intensive. We trained and evaluated the model on an NVIDIA T4 GPU (16 GB VRAM) and assessed inference latency on an NVIDIA GTX 980M (4 GB VRAM). The resulting per-image inference time is 966 ms (~ 1.0 FPS). This performance suggests feasible interactive inference on legacy, mobile-class GPU and indicates potential for edge deployment on resource-constrained devices. In parallel, cloud-based GPUs can support a scalable, high-throughput monitoring workflow, making the proposed approach suitable for large-scale operational agricultural disease surveillance.

## Performance Metrics

1. ***True Positive (TP)*** - A true positive occurs when a positive instance is correctly predicted as positive by the model. It represents correctly identified positive cases [51] .
2. ***True Negative (TN)*** - A true negative occurs when a negative instance is correctly predicted as negative by the model. It reflects correctly identified negative cases [51].
3. ***False Positive (FP)*** - A false positive occurs when a negative instance is incorrectly predicted as positive by the model. This error is often associated with the false alarm rate [51].
4. ***False Negative (FN)*** - A false negative occurs when a positive instance is incorrectly predicted as negative by the model. This error is commonly referred to as the miss rate or under-reporting rate [51].
5. ***Accuracy*** - This refers to the weighting of the correct decision by the classifier. Range [0,1]. Direction: higher is better [51].

$$A = \frac{\sum TP+T}{\sum TP+TN+FP+FN} \quad (8)$$

6. ***Precision*** - Precision measures the proportion of true positive predictions among all positive predictions made by a model. Range [0,1]. Direction: higher is better [51].

$$P = \frac{\sum TP}{\sum TP+FP} \quad (9)$$

7. ***Recall*** - Recall measures the proportion of true positive predictions among all actual positive instances in the dataset. Range [0,1]. Direction: higher is better [51].

$$R = \frac{\sum TP}{\sum TP+F} \qquad (10)$$

8. ***F1-score*** - Provides a single value that balances precision and recall, and is often used as an alternative metric in scenarios where a unified view of detection performance is desired [51].

$$F1\ Score = 2 * \frac{Precision*Recall}{Precision+Recall} \qquad (11)$$

TABLE IV. DEEP LEARNING MODELS COMPARISON

| ***Models*** | ***Size MB*** | ***Para meter (Milli ons)*** | ***Depth*** | ***Input Image Sizes*** | ***Avg. Accur acy (%) [45]*** |
|---|---|---|---|---|---|
| DenseNet201 [52] | 77.4 | 20.1 | 201 | 224x224 | 93.48 |
| ResNet101 [52] | 171 | 44.6 | 101 | 224x224 | 93.25 |
| ResNet50 [52] | 97.8 | 25.6 | 50 | 224x224 | 93.03 |
| GoogLeNet [52] | 49.7 | 6.62 | 22 | 224x224 | 87.62 |
| AlexNet [52] | 233 | 61.1 | 8 | 224x224 | 86.93 |
| ResNet18 [52] | 44.7 | 11.7 | 18 | 224x224 | 91.45 |
| EfficientNet-B0 [52] | 21 | 5.3 | 237 | 224x224 | 93.16 |
| NASNetMob [52] | 20 | 5.3 | ~ | 224x224 | 92.6 |
| ConvNeXtS [52] | 192 | 50.2 | ~ | 224x224 | 92.90 |
| **Our** | **14.3** | 1.251 | **18** | **224x224** | **99.53** |

Table IV reports a benchmark against trending pre-trained models for plant disease detection cases [45]. As shown, PD36-C achieves competitive performance, attributable to an improved training strategy and a more comprehensive training set, with notably: ~1.251 M parameters, ~14.2 MB model size, 18-layer depth, and average test accuracy 99.53%. Evaluation set metrics for training reached best validation/test accuracy at epoch 29 (0.9953) and best validation/test loss at epoch 30 (0.4290). On the test set, balanced accuracy reached (0.9952), Mathews Correlation Coefficient (0.9952), Cohen's Kappa (0.9952), and Macro-average AUC (0.9999).

TABLE V. TRAINING HISTORY

| ***Ep och*** | ***Learnin g rate*** | ***Training accuracy*** | ***Training loss*** | ***Validation accuracy*** | ***Validation loss*** |
|---|---|---|---|---|---|
| 1 | 0.0001 | 0.61815 | 1.61720 | 0.56442 | 1.71441 |
| 2 | 0.0001 | 0.84631 | 0.92046 | 0.63237 | 1.52438 |
| 4 | 0.0001 | 0.91082 | 0.74684 | 0.84635 | 0.90112 |
| 4 | 0.0001 | 0.93741 | 0.66958 | 0.90752 | 0.73396 |
| 5 | 0.0001 | 0.95261 | 0.62247 | 0.90752 | 0.72019 |
| 6 | 0.0001 | 0.96170 | 0.59189 | 0.95077 | 0.61142 |
| 7 | 0.0001 | 0.96906 | 0.56933 | 0.95573 | 0.59623 |
| 8 | 0.0001 | 0.97384 | 0.55181 | 0.95669 | 0.59102 |
| 9 | 0.0001 | 0.97747 | 0.53879 | 0.96409 | 0.56241 |
| 10 | 0.0001 | 0.97980 | 0.52907 | 0.96705 | 0.54706 |
| 11 | 0.0001 | 0.98173 | 0.51784 | 0.90513 | 0.72912 |
| 12 | 0.0001 | 0.98364 | 0.51113 | 0.97234 | 0.53607 |
| 13 | 0.0001 | 0.98607 | 0.50197 | 0.98139 | 0.50151 |
| 14 | 0.0001 | 0.98782 | 0.49516 | 0.97388 | 0.53411 |
| 15 | 0.0001 | 0.98835 | 0.48976 | 0.97069 | 0.52888 |
| 16 | 0.00005 | 0.99327 | 0.46638 | 0.99141 | 0.45469 |
| 17 | 0.00005 | 0.99398 | 0.46214 | 0.99289 | 0.44672 |
| 18 | 0.00005 | 0.99435 | 0.45969 | 0.99397 | 0.44387 |
| 19 | 0.00005 | 0.99499 | 0.45706 | 0.99380 | 0.44303 |
| 20 | 0.00005 | 0.99495 | 0.45507 | 0.98464 | 0.47207 |
| 21 | 0.00005 | 0.99495 | 0.45512 | 0.98976 | 0.45508 |
| 22 | 0.00005 | 0.99553 | 0.45321 | 0.99283 | 0.44060 |
| 23 | 0.00005 | 0.99579 | 0.45141 | 0.99351 | 0.44408 |
| 24 | 0.00005 | 0.99553 | 0.45044 | 0.99368 | 0.43941 |
| 25 | 0.00005 | 0.99626 | 0.44829 | 0.99391 | 0.43781 |
| 26 | 0.00005 | 0.99607 | 0.44758 | 0.99402 | 0.43571 |
| 27 | 0.00005 | 0.99659 | 0.44525 | 0.99397 | 0.43370 |
| 28 | 0.00005 | 0.99659 | 0.44444 | 0.99442 | 0.43172 |
| 29 | 0.00005 | 0.99673 | 0.44320 | 0.99420 | 0.43078 |
| 30 | 0.00005 | 0.99697 | 0.44110 | 0.99533 | 0.42897 |

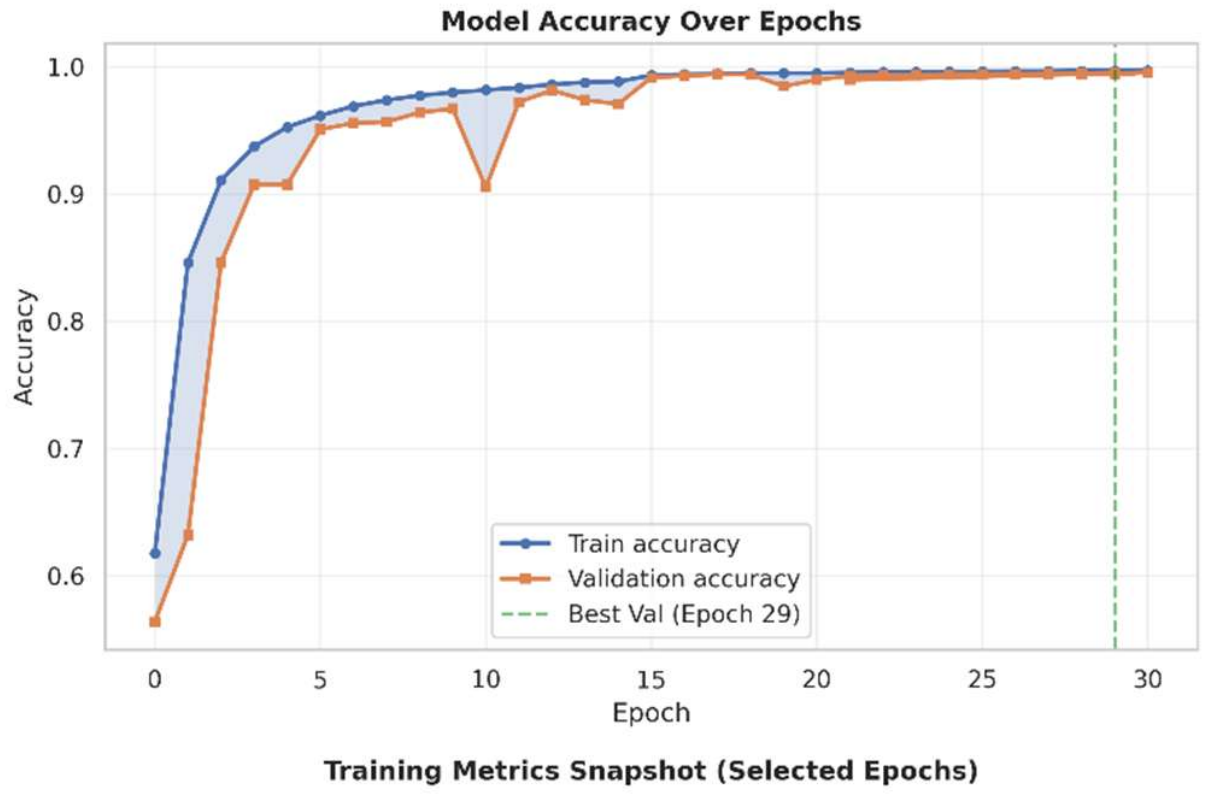

Fig. 12 Training History of PD36-C

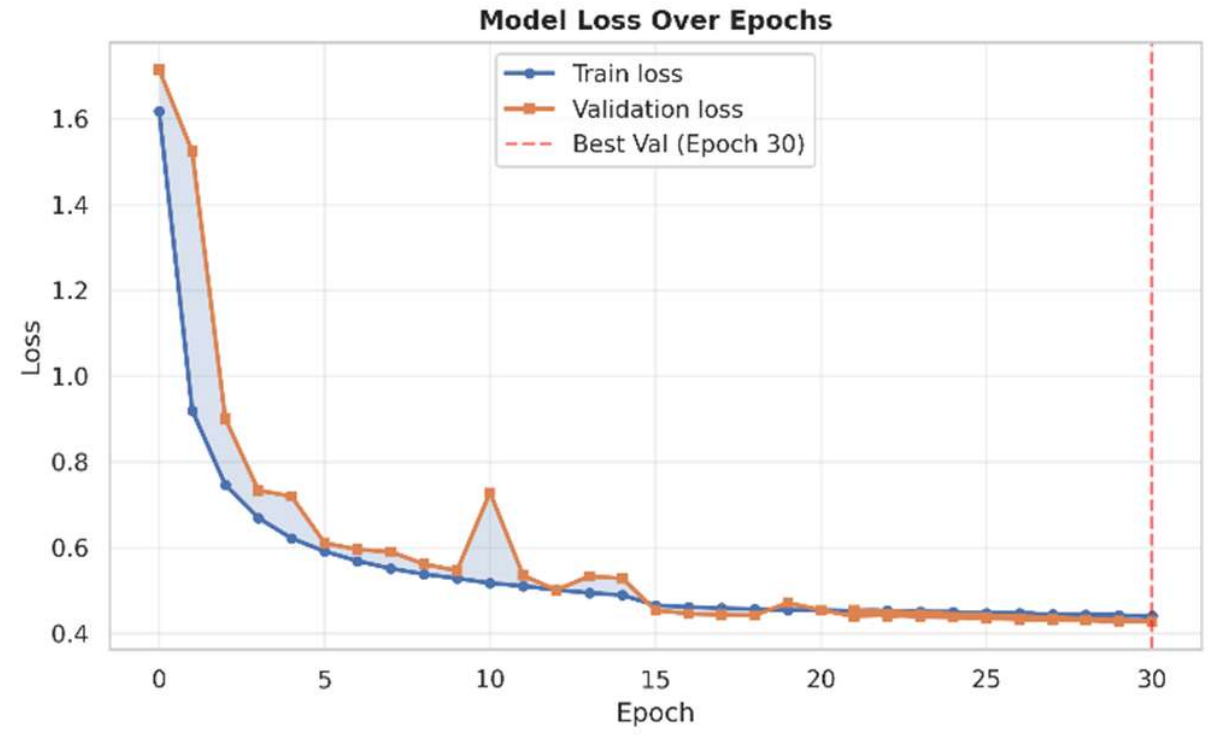

Fig. 13 Training History of PD36-C

Table V summarizes the training history over thirty epochs at a learning rate of $1 \times 10^{-4}$ to $0.5 \times 10^{-5}$, with Fig. 12 and Fig. 13 visualizing the trajectory. The model initializes at modest performance in Epoch 1 (training accuracy-TA = 0.6182, training loss-TL = 1.6172, validation accuracy-VA 0.5644, validation loss-VL 1.7144), as expected from random initialization and limited early optimization. By Epoch 2, performance improves (TA = 0.8463, TL = 0.9205, VA 0.6324, VL 1.5244), followed by continued gains in Epoch 3 (TA = 0.9108, TL = 0.7468, VA 0.8464, VL 0.9011). From Epochs 4-10, the model continues improvement, culminating in (TA = 0.9798, TL = 0.5291, VA 0.9671, VL 0.5471). While from Epochs 16-30, the trends remain monotonic and stable. The peak accuracy is reached at epoch 30 with (TA = 0.99697, TL = 0.4411, VA 0.9953, VL 0.42897). Taken together, these curves in Fig. 12 and Fig. 13 exhibit stable convergence without uncontrolled oscillations, indicating effective optimization and model capacity. The modest gap between the final training accuracy and the test accuracy aligns with reasonable generalization given the model's size and the variability inherent in the data.

### 4.1 Confusion Matrix Analysis

We evaluate detection performance across 38 classes exhibiting diverse morphology, texture, and color patterns using per-class precision, recall, and F1-score (Table VI). This analysis complements aggregate metrics by revealing class-specific strengths and failures.

TABLE VI. MODEL CLASSIFICATION METRICS

| *Nr* | *class* | *precision* | *recall* | *f1_score* |
|---|---|---|---|---|
| 0 | Apple Apple_scab | 0.99802 | 1.00000 | 0.99901 |
| 1 | Apple Black_rot | 1.00000 | 1.00000 | 1.00000 |
| 2 | Apple Cedar_apple_rust | 1.00000 | 1.00000 | 1.00000 |
| 3 | Apple healthy | 0.99603 | 1.00000 | 0.99801 |
| 4 | Blueberry healthy | 1.00000 | 1.00000 | 1.00000 |
| 5 | Cherry_(including_sour) Powdery_mildew | 1.00000 | 1.00000 | 1.00000 |
| 6 | Cherry_(including_sour) healthy | 1.00000 | 1.00000 | 1.00000 |
| 7 | Corn_(maize) Cercospora_leaf_spot Gray_leaf_spot | 0.97772 | 0.96341 | 0.97052 |
| 8 | Corn_(maize) Common_rust_ | 0.99790 | 0.99581 | 0.99685 |
| 9 | Corn_(maize) Northern_Leaf_Blight | 0.96488 | 0.97904 | 0.97190 |
| 10 | Corn_(maize) healthy | 0.99785 | 1.00000 | 0.99893 |
| 11 | Grape Black_rot | 1.00000 | 1.00000 | 1.00000 |
| 12 | Grape Esca_(Black_Measles) | 1.00000 | 1.00000 | 1.00000 |
| 13 | Grape Leaf_blight (Isariopsis_Leaf_Spot) | 1.00000 | 1.00000 | 1.00000 |
| 14 | Grape healthy | 1.00000 | 1.00000 | 1.00000 |
| 15 | Orange Huanglongbing (Citrus_greening) | 0.99604 | 1.00000 | 0.99802 |
| 16 | Peach Bacterial_spot | 1.00000 | 0.99782 | 0.99891 |
| 17 | Peach healthy | 0.99538 | 0.99769 | 0.99653 |
| 18 | Pepper_bell Bacterial_spot | 1.00000 | 1.00000 | 1.00000 |
| 19 | Pepper_bell healthy | 0.98413 | 0.99799 | 0.99101 |
| 20 | Potato Early_blight | 1.00000 | 1.00000 | 1.00000 |
| 21 | Potato Late_blight | 0.99384 | 0.99794 | 0.99588 |
| 22 | Potato healthy | 0.99777 | 0.98246 | 0.99006 |
| 23 | Raspberry healthy | 1.00000 | 0.99775 | 0.99888 |
| 24 | Soybean healthy | 0.99604 | 0.99604 | 0.99604 |
| 25 | Squash Powdery_mildew | 1.00000 | 0.99078 | 0.99537 |
| 26 | Strawberry Leaf_scorch | 1.00000 | 1.00000 | 1.00000 |
| 27 | Strawberry healthy | 1.00000 | 1.00000 | 1.00000 |
| 28 | Tomato Bacterial_spot | 0.98829 | 0.99294 | 0.99061 |
| 29 | Tomato Early_blight | 0.99789 | 0.98333 | 0.99056 |
| 30 | Tomato Late_blight | 0.98077 | 0.99136 | 0.98604 |
| 31 | Tomato Leaf_Mold | 1.00000 | 0.99787 | 0.99894 |
| 32 | Tomato Septoria_leaf_spot | 0.99770 | 0.99312 | 0.99540 |
| 33 | Tomato Spider_mites Two-spotted_spider_mite | 0.99074 | 0.98391 | 0.98731 |
| 34 | Tomato Target_Spot | 0.98048 | 0.98906 | 0.98475 |
| 35 | TomatoYellow_Leaf_Curl_Virus | 1.00000 | 0.98980 | 0.99487 |
| 36 | Tomato_mosaic_virus | 0.99777 | 1.00000 | 0.99889 |
| 37 | Tomato healthy | 0.99380 | 1.00000 | 0.99689 |
| | accuracy | | | 0.99533 |
| | macro avg | 0.99534 | 0.99521 | 0.99527 |
| | weighted avg | 0.99536 | 0.99533 | 0.99533 |

As reported in Table VI and visualized in Fig. 14, the model attains high detection accuracy across most categories. Per-class performance spans: Lower end: Corn (maize) - Cercospora leaf spot/Gray leaf spot with precision ~ 0.9777 and recall ~ 0.9634, indicating occasional confusion with visually similar categories. Upper end: numerous classes, including Apple Black rot, Cedar apple rust, Blueberry healthy, Cherry Powdery mildew, Cherry healthy, and all four Grape categories, achieve perfect precision and recall of 1.00, reflecting highly distinctive symptomatology and cleaner images.

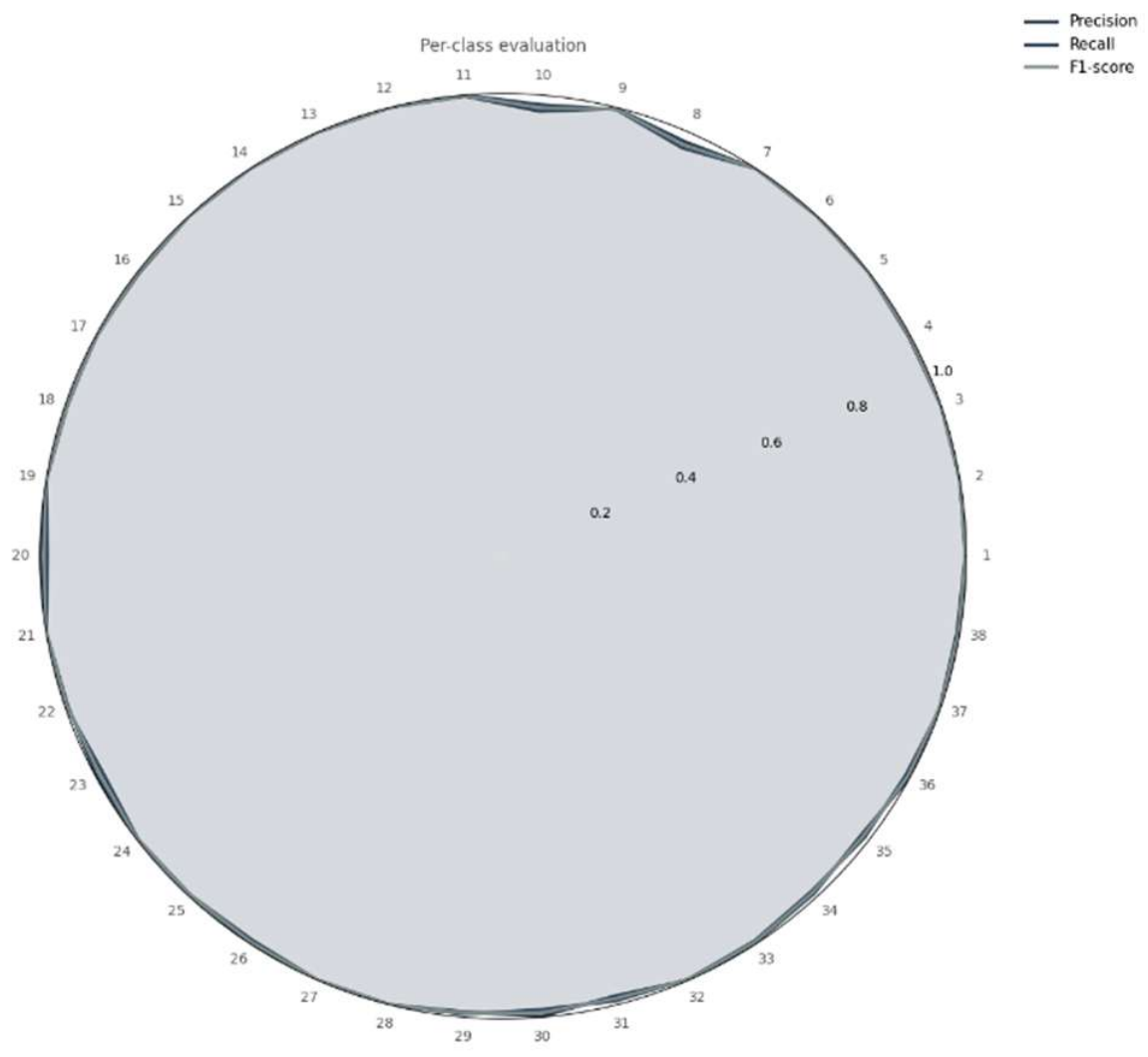

Fig. 14 Precision, recall, and F1 score for each class.

These extremes are consistent with feature separability: classes with pronounced, high-contrast lesions and distinctive morphology (e.g., powdery mildew on cherry, grape diseases) are more easily detected, while subtle, low-contrast symptoms (e.g., some maize foliar diseases) reduce separability and increase confusion. In such cases, CNNs may struggle to extract robust features from low-signal regions, particularly under variable illumination or motion blur.

The confusion matrix for the validation partition, Fig. 15, shows diagonal dominance, with per-class true positive rates generally high. The worst performing class is Corn (maize) Cercospora leaf spot / Gray leaf spot, achieving a recall of 96.34%, while the majority of classes reach 100% on both precision and recall. Overall accuracy on this evaluation set is 99.53%. These results align with prior reports on CNN-based leaf diagnostics [14]. As expected, when evaluating with web images captured outside controlled/studio conditions, several studies report a drop in accuracy due to domain shift and background clutter [53], [44].

### 4.2 Gradient-weighted Class Activation Mapping(Grad-CAM)

To interpret individual predictions, we implement Grad-CAM. A functional sub-model is constructed from the network input to the final Conv2D layer and the softmax output. For a given input image and target class index c, the gradient of the class score $y^c$ with respect to every spatial activation in the final convolutional feature map is computed via a single forward pass. The gradients are global-average-pooled over the spatial dimensions to obtain per-channel importance weights, which are then used to form a weighted linear combination of the activation maps. A ReLU is applied to retain only positively contributing activations, and the resulting heatmap is min-max normalized [0,1] and bilinearly unsampled to $224 \times 224$.

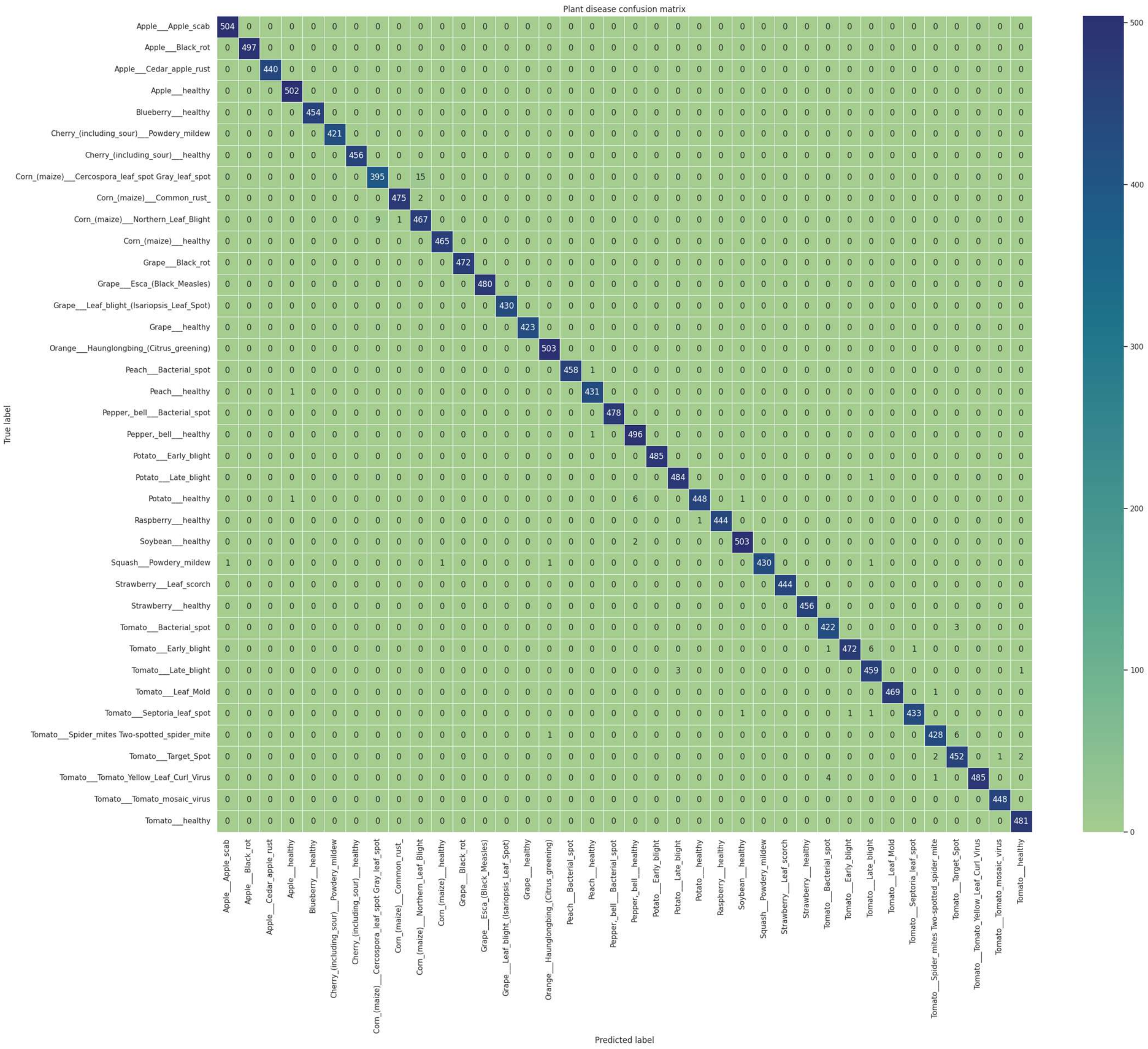

Fig. 15 Confusion Matrix for the 38-class plant diseases. Rows represent ground-truth classes and columns represent predicted classes; diagonal cells indicate correct classifications.

In Fig. 16, predicted class labels are highlighted in red to indicate errors. Compared with correctly classified cases, misclassified samples exhibit either diffuse, low-contrast activation spread across the entire leaf surface, suggesting insufficient discriminative signals or attention concentrated on a region that overlaps visually similar symptoms. These cases reveal failure of the model and suggest that diseases with morphologically similar lesion phenotypes, such as early blight and target spot in tomato, or gray leaf spot and northern leaf blight in maize, represent the most challenging classes.

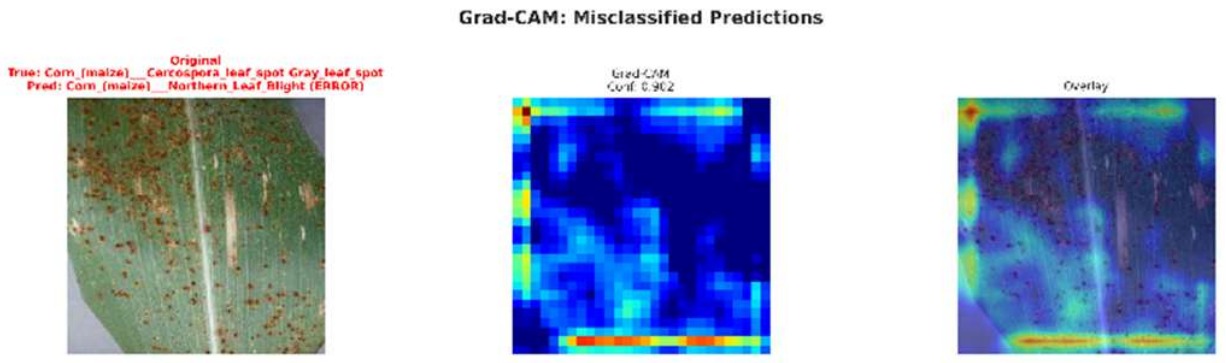

Fig. 16 Gradient-weighted Class Activation Maps (Grad-CAM) analysis of misclassified test sample.

In Fig. 17 each row presents one sample across three columns: (left) the original RGB image with ground-truth and predicted class label, (middle) the raw Grad-CAM heatmap produced from the final convolutional layer, rendered on a colormap where red/yellow indicate high gradient weight and blue indicate area with low interest, (right) the heatmap is blended with original image at $x = 0.5$ opacity. In all four samples, the network attention is concentrated on the visible lesion region (discolored patches), specific spots or fungal growth spots, rather than on the background, confirming that the classifier's decisions are grounded in pathologically relevant visual features.

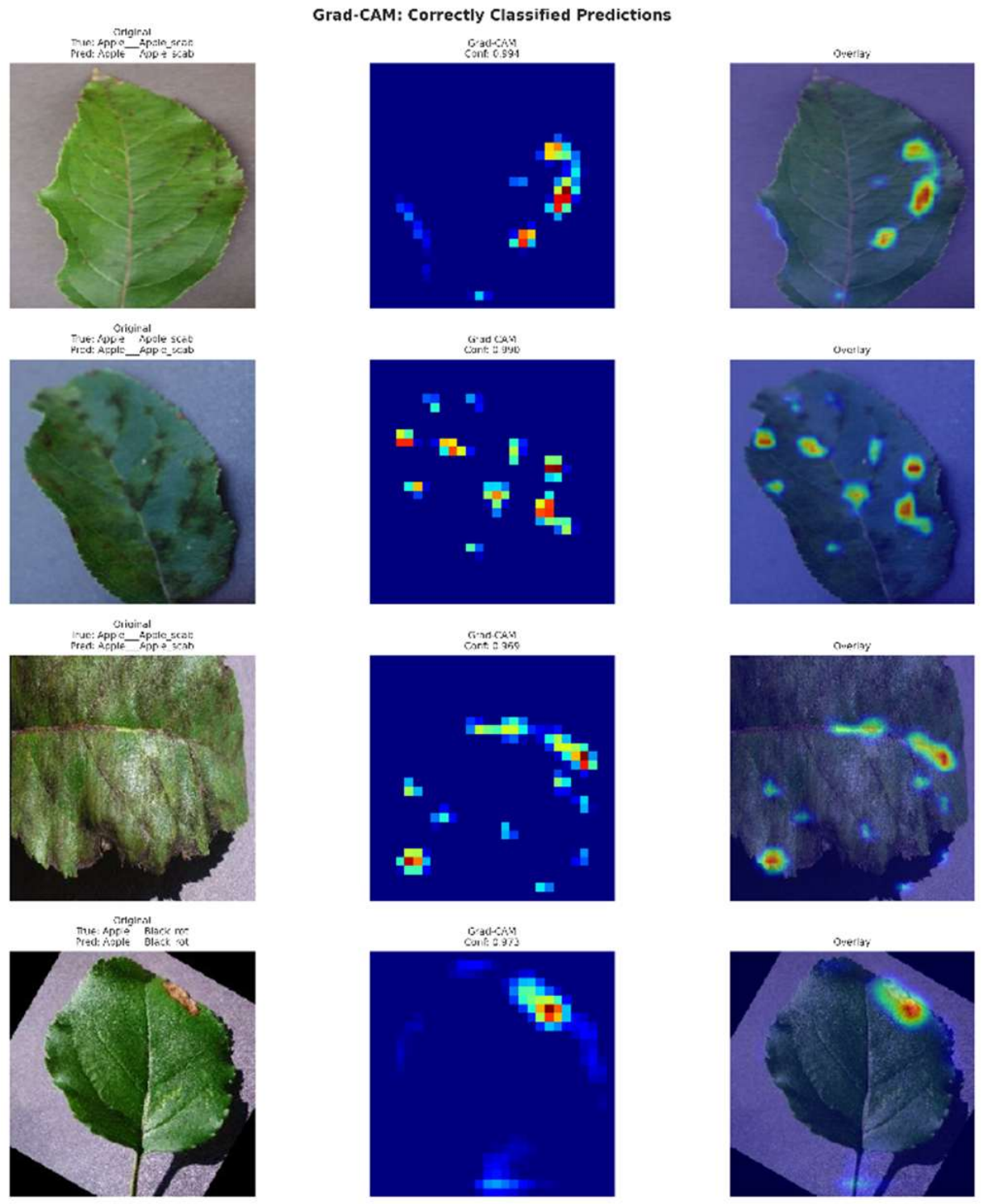

Fig. 17 Gradient-weighted Class Activation Maps (Grad-CAM) for four correctly classified test samples.

## 4.3 C_Best, C_Second_Best and C_True

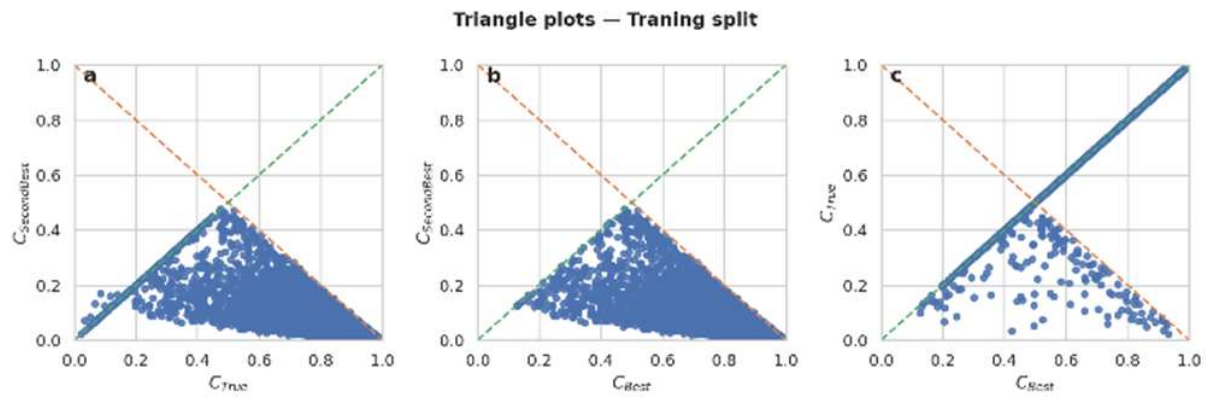

Fig. 18 Scatter plot showing confidence for the true class ($C_{True}$) or Best class $C_{Best}$ compared to the confidence of the second-best choice ($C_{SecondBest}$) on the training dataset.

To test model clarity and ambiguity, we analyze the top-k probability shown in Fig. 18, Fig. 19, and Fig. 20. It shows the scatter of ( $C_{\text{True}}, C_{\text{Best}}, C_{\text{SecondBest}}$ ), where $C_{\text{True}}$ is the probability assigned to the ground-truth class, $C_{\text{Best}}$ is the first-highest and $C_{\text{SecondBest}}$ is the second-highest predicted probability. The red line $C_{\text{True}} + C_{\text{SecondBest}} = 1$ marks cases where nearly all probability mass concentrates in the top-2 classes. Points near the axes (high $C_{\text{SecondBest}}$ , low $C_{\text{True}}$ ) Indicate confident misclassifications, diffuse points below the red line suggest broader uncertainty spread across more than two classes. Distribution of the confidence margin $C_{\text{True}} - C_{\text{SecondBest}} = 0$. Points near zero (the green line) denote ambiguous cases where the top-2 classes are nearly indistinguishable, large positive margins indicate confident correct predictions, and negative margins flag errors. Points along the diagonal $C_{\text{True}} = C_{\text{Best}}$ correspond to correct top1 predictions, proximity to the red boundary $C_{\text{True}} + C_{\text{Best}} = 1$ indicates that the top class gets nearly all attention when correct, whereas deviations highlight uncertainty or class competition.

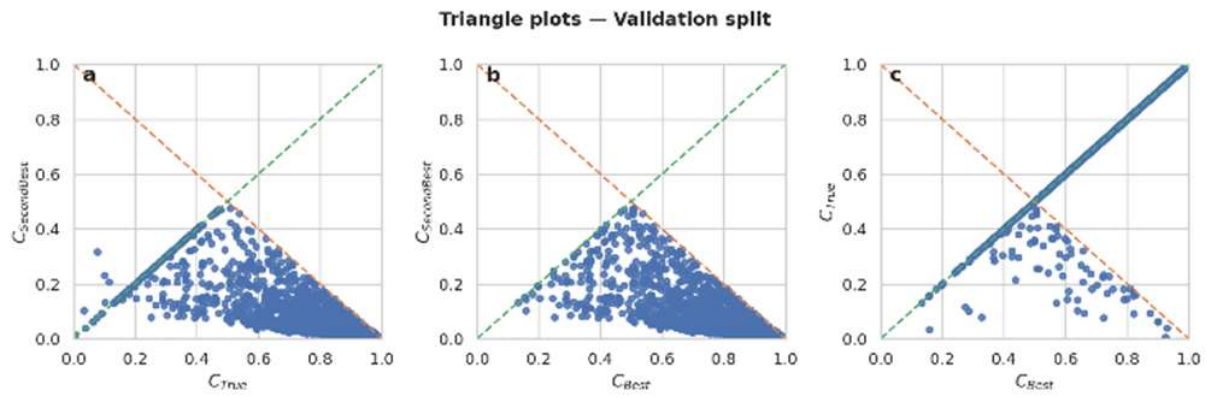

Fig. 19 Scatter plot showing confidence for the true class ($C_{True}$) or Best class $C_{Best}$ compared to the confidence of the second-best choice ($C_{SecondBest}$) on the validation dataset.

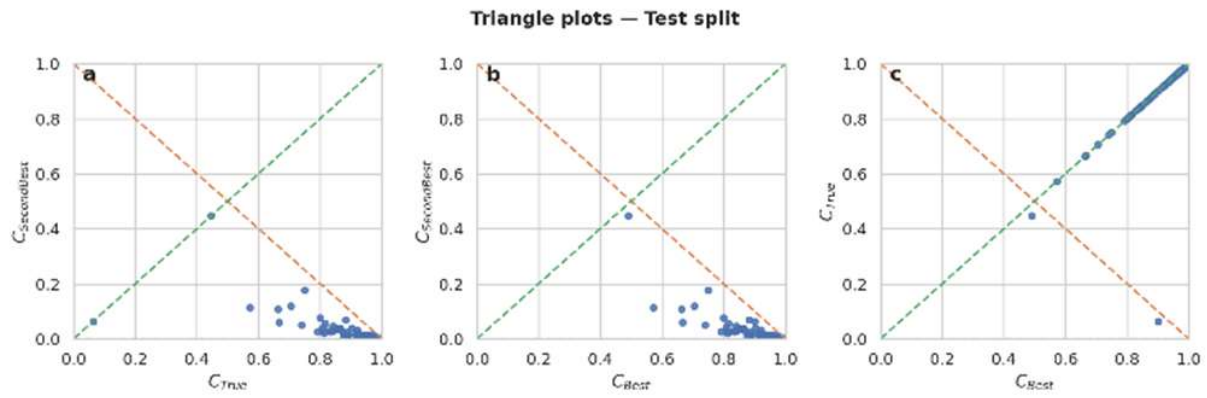

Fig. 20 Scatter plot showing confidence for the true class ($C_{True}$) or Best class $C_{Best}$ compared to the confidence of the second-best choice ($C_{SecondBest}$) on the test dataset.

## V. APP DESIGN

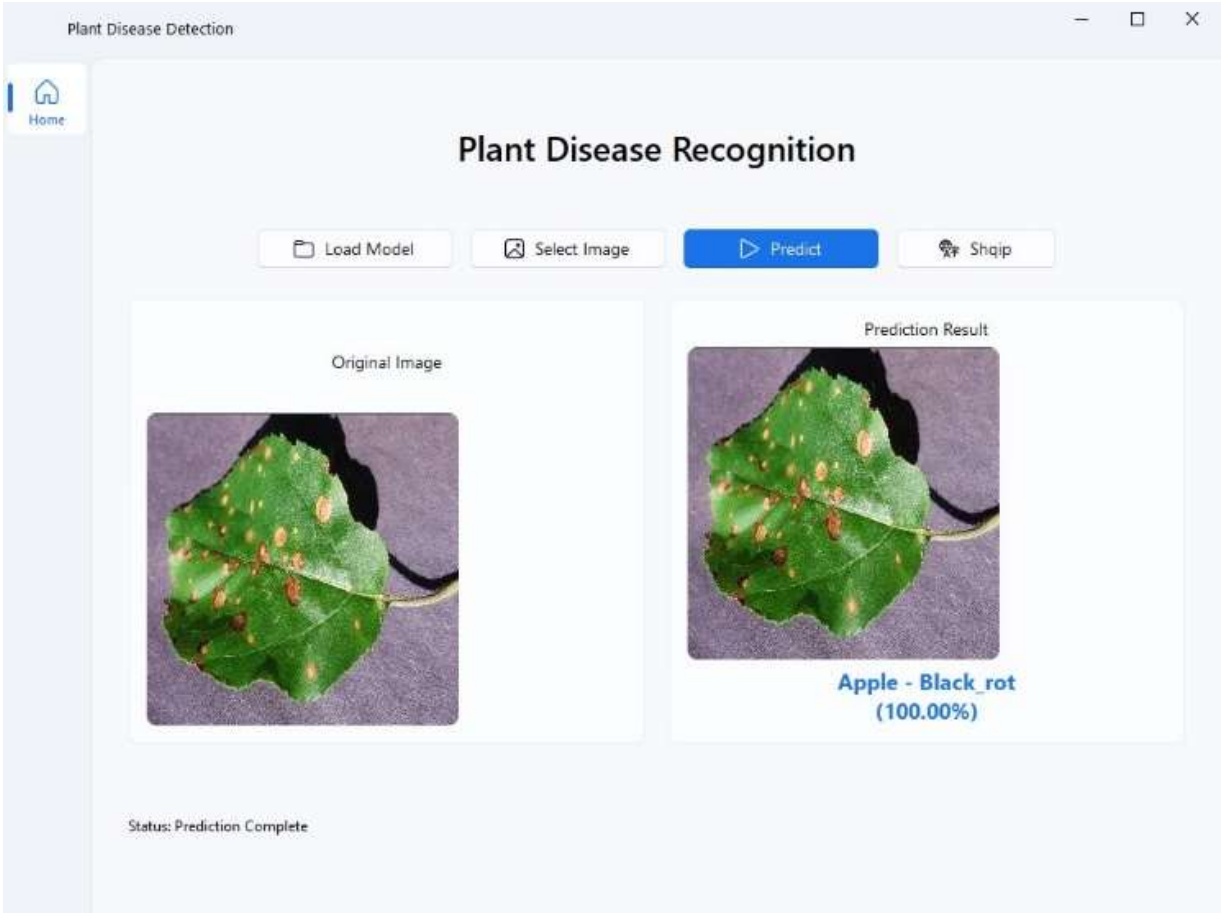

Fig. 21 Application UI

The graphical user interface (GUI) was implemented with the Qt framework for Python, delivering a clean, user-friendly Material design aesthetic aligned with the Windows 11 design style Fig. 21. Adherence to Material Design principles ensures visual clarity, consistency across components, and intuitive interaction patterns suitable for plant-disease detection workflows. Given the model's strong detection performance and compact footprint, ~1.251 million parameters, and 14.3 MB on disk, a Windows desktop application was developed to enable practical testing and usage. The application allows users to load images from local storage and perform real-time inference. As illustrated in Fig. 22, the system successfully identifies Apple-Black rot on an input image with 100% confidence and subsequently triggers a pop-up dialog that provides a concise disease description and treatment recommendations. The application layout comprises five primary areas (Fig. 22): (a) a *Tab Bar*, (b) a *Control Button Bar* with Load Model (select a model), Select Image (choose an input image), and Predict (run inference with PD36-C or other models), (c) *Original and Predicted Image* viewports displaying the predicted class label and confidence (accuracy percentage), (d) a *Status Bar*, and (e) an *Application Title Bar*.

## VI. Discussion

This study addresses plant disease detection in agriculture using ML and CNN-based models. We first conducted a detailed review of existing studies, surveying techniques, datasets, and model architectures. Building on this analysis, we proposed a compact model (PD36-C) that achieves an average accuracy of approximately 99.53%. Furthermore, our review confirms that contemporary CNN approaches routinely exceed 90% accuracy in this domain, underscoring their suitability for plant disease recognition [32], [42], [45], [54], [55], [56], [57].

The literature indicates that architectural transformations such as increasing depth, widening layers, or adding specialized modules can improve accuracy, at the cost of higher computational demands and longer training times. Recently, vision transformers have been proposed to enhance feature extraction and robustness under challenging conditions. Performance can also be improved by using diverse input modalities or hybrid methods (e.g., incorporating time-series signals), which can enrich contextual information and stabilize predictions [15], [18], [32], [38], [40], [43], [45], [50], [57], [58], [59], [60], [61], [62], [63], [64], [65].

Among numerous studies, several top-performing models are widely adopted and repeatedly benchmarked (Table IV) [40], [50], [58], [59], [62], [63], [65]. These models can be used off-the-shelf in new scenarios or fine-tuned with additional training to address domain shifts [18], [38], [43], [45], [57], [60]. Notably, some studies implement models from scratch [42], [15], [18], [32], [40], [61], [64]. While some works rely on established datasets, others curate new datasets. However, such datasets often remain limited in diversity and size [18], [64].

In the following, we address the research questions:

**RQ1**: Why is plant disease detection of significant importance in the area of agriculture?

Agriculture is a cornerstone of global food security, and plant diseases can substantially degrade both the quality and quantity of crops. Early, accurate detection is therefore essential to mitigate losses, guide timely interventions, and protect yields [11], [10], [15], [27], [66], [42], [14], [11], [18], [26], [55].

**RQ2**: To what extent can DL models address key challenges in plant disease detection, and what approaches yield efficient, compact ("***tiny***") models?

DL, particularly CNNs are the prevailing paradigm for image classification. By learning hierarchical visual features, CNNs can solve complex recognition tasks with high efficiency and low error rates, achieving strong predictive accuracy for plant diseases with visible symptoms [43], [29], [38], [10], [54], [27], [42], [14], [11]. Efficient, compact architectures (e.g., carefully pruned CNNs, depthwise-separable convolutions), compared to transfer learning, enable high performance under constrained resources.

**RQ3**: Which model architectures are most suitable for accurate, robust foliar disease recognition?

Architecture selection depends on disease characteristics, dataset properties, hardware constraints, and target accuracy. In most settings, CNNs deliver strong performance (e.g., our PD36-C achieves ~99.53% average accuracy across 38 classes during training) while RNNs are more specialized for temporal signals [11]. Current practice favors transfer learning and fine-tuning with additional data, and, where beneficial, integrating more advanced modules to enhance representation capacity [18], [50], [58], [59], [61], [62], [63], [64], [65].

**RQ4**: Can DL models operate reliably on resource-constrained edge devices in offline settings?

Accuracy remains the most commonly reported metric, reflecting the end-to-end capability for segmentation, feature extraction, and correct classification. With many CNNs exceeding 90% accuracy, these models demonstrate reliability and the potential to outperform traditional methods in timely disease identification. With appropriate optimization, they can be adapted for offline, edge deployment [32], [42], [45], [54], [55], [56], [57].

**RQ5**: How do ML/DL methods compare with traditional diagnostic practices in accuracy and reliability?

Lightweight fine-tuned DNNs (e.g., MobileNetV2) and compact models such as our PD36-C can deliver high accuracy with only a few million parameters and modest depth, reducing computational load and enabling practical use on commodity mobile devices. These capabilities complement traditional manual diagnostics in speed, consistency, and scalability [62], [43], [35], [60], [45].

**RQ6**: Do current models produce stable predictions across diverse crops, diseases, and environments, and what are the main limitations?

Current models tend to perform well withing the narrow conditions they were trained on, but lack the robustness needed for reliable deployment across the full spectrum of crops, disease types, and real-world environments. Key limitations include dependence on large labeled datasets, substantial computing for training/deployment, risk of overfitting with limited or imbalanced data, and sensitivity to data quality, symptom variability, and domain shift. Models may struggle to localize disease regions, to detect multiple co-occurring diseases on the same leaf, or to integrate with operational systems [48], [14], [11], [18].

### A. Limitations and Future Work

Despite the high accuracy of the model, the current study also acknowledges several limitations:

- *First*, near-real-time performance remains insufficient for high-throughput pipelines and low-power edge devices. Extending to multi-camera, 24/7 monitoring at farm/region scale strains compute, storage, and bandwidth challenges.
- *Second*, intermittent power and environmental artifacts (dust, motion, bird occlusions) increase misclassification risk. Background clutter, noise, and non-studio conditions degrade feature extraction and generalization.
- *Finally*, co-occurring or look-alike symptoms and very early lesions remain challenging.

Additionally, a comparative analysis should be conducted to benchmark the proposed model against future research, which can further refine the model performance:

- *Transfer/self-supervised learning*: Fine-tuning PD36-C and exploiting unlabeled data to improve robustness on rare and shifted classes.

- *Multimodal & temporal cues*: Integrating hyperspectral/thermal/multispectral inputs and short time-series to boost early-stage detection.
- *Edge optimization & data strategy*: Applying pruning, quantization, and knowledge distillation to cut latency and memory for offline edge deployment.

Addressing these limitations, future recommendations will be critical for transitioning from a prototype to a production-ready system capable of supporting smart agriculture.

## VII. Conclusion

Plant diseases and pests pose a significant threat to global agricultural production. Contemporary detection still relies heavily on manual inspection and expert judgment, which are inefficient and not reliably scalable to large-area deployments, while laboratory examinations, though accurate, are expensive and time-consuming. To address these limitations, this study proposes a tiny CNN for leaf-based disease detection, accompanied by an application for edge-device testing and use. Empirical results indicate that optimized deep CNNs can achieve strong performance on large, diverse datasets, making them both effective and efficient for practical deployment. The primary contribution of this work lies in developing a compact, efficient model that is ready for real-world detection on devices with limited hardware resources. The confusion matrix (Fig. 15) shows strong diagonal dominance, with most errors arising among visually similar classes, consistent with fine-grained pathology. Finally, an agronomic interpretation module augments the classifier by providing concise disease analyses and decision support for practitioners. Overall, the system attains an accuracy of 99.53% across all dataset categories.

## VIII. Declarations

**Conflict of Interest**: The authors declare that they have no conflict of interest.

**Funding**: No funding was received for this study.

**Ethical Approval**: Not applicable.

**Consent to Participate**: Not applicable.

**Consent for Publication**: Not applicable.

## References

[1] S. Yadav, N. Sengar, A. Singh, A. Singh, and M. K. Dutta, "Identification of disease using deep learning and evaluation of bacteriosis in peach leaf," *Ecol. Inform.*, vol. 61, p. 101247, Mar. 2021, doi: 10.1016/j.ecoinf.2021.101247.

[2] K. S. Poutanen *et al.*, "Grains – a major source of sustainable protein for health," *Nutr. Rev.*, vol. 80, no. 6, pp. 1648–1663, Jun. 2022, doi: 10.1093/nutrit/nuab084.

[3] S. SHERIFI, S. ISMAILI, F. IDRIZI, and E. RUSTEMI, "ALBANK - A CASE STUDY ON THE USE OF ETHEREUM BLOCKCHAIN TECHNOLOGY AND SMART CONTRACTS FOR SECURE DECENTRALIZED BANK APPLICATION," *J. Nat. Sci. Math. UT*, vol. 10, no. 19–20, pp. 380–400, Dec. 2025.

[4] S. Sherifi, F. Halili, and M. Kasa-Halili, "Intelligent Traffic Monitoring with YOLOv11: A Case Study in Real-Time Vehicle Detection," in *2025 International Conference on Computer and Applications (ICCA)*, Dec. 2025, pp. 1–7. doi: 10.1109/ICCA66035.2025.11430921.

[5] G. Han *et al.*, "Identification of an Elite Wheat-Rye T1RS·1BL Translocation Line Conferring High Resistance to Powdery Mildew and Stripe Rust," *Plant Dis.*, vol. 104, no. 11, pp. 2940–2948, Nov. 2020, doi: 10.1094/PDIS-02-20-0323-RE.

[6] S. Nigam *et al.*, "Deep transfer learning model for disease identification in wheat crop," *Ecol. Inform.*, vol. 75, p. 102068, Jul. 2023, doi: 10.1016/j.ecoinf.2023.102068.

[7] S. K.m., S. V., S. K. P., and S. O.k., "AI based rice leaf disease identification enhanced by Dynamic Mode Decomposition," *Eng. Appl. Artif. Intell.*, vol. 120, p. 105836, Apr. 2023, doi: 10.1016/j.engappai.2023.105836.

[8] Md. A. Haque, S. Marwaha, C. K. Deb, S. Nigam, and A. Arora, "Recognition of diseases of maize crop using deep learning models," *Neural Comput. Appl.*, vol. 35, no. 10, pp. 7407–7421, Apr. 2023, doi: 10.1007/s00521-022-08003-9.

[9] H. Yu *et al.*, "Corn Leaf Diseases Diagnosis Based on K-Means Clustering and Deep Learning," *IEEE Access*, vol. 9, pp. 143824–143835, 2021, doi: 10.1109/ACCESS.2021.3120379.

[10] C. Jackulin and S. Murugavalli, "A comprehensive review on detection of plant disease using machine learning and deep learning approaches," *Meas. Sens.*, vol. 24, p. 100441, Dec. 2022, doi: 10.1016/j.measen.2022.100441.

[11] A. Singla *et al.*, "Exploration of machine learning approaches for automated crop disease detection," *Curr. Plant Biol.*, vol. 40, p. 100382, Dec. 2024, doi: 10.1016/j.cpb.2024.100382.

[12] S. Kaur, S. Pandey, and S. Goel, "Plants Disease Identification and Classification Through Leaf Images: A Survey," *Arch. Comput. Methods Eng.*, vol. 26, no. 2, pp. 507–530, Apr. 2019, doi: 10.1007/s11831-018-9255-6.

[13] P. Sahu, A. Chug, A. P. Singh, D. Singh, and R. P. Singh, "Challenges and Issues in Plant Disease Detection Using Deep Learning:," M. Dua and A. K. Jain, Eds., IGI Global, 2021, pp. 56–74. doi: 10.4018/978-1-7998-3299-7.ch004.

[14] A. Upadhyay *et al.*, "Deep learning and computer vision in plant disease detection: a comprehensive review of techniques, models, and trends in precision agriculture," *Artif. Intell. Rev.*, vol. 58, no. 3, p. 92, Jan. 2025, doi: 10.1007/s10462-024-11100-x.

[15] H. Li, L. Huang, C. Ruan, W. Huang, C. Wang, and J. Zhao, "A dual-branch neural network for crop disease recognition by integrating frequency domain and spatial domain information," *Comput. Electron. Agric.*, vol. 219, p. 108843, Apr. 2024, doi: 10.1016/j.compag.2024.108843.

[16] V. Bischoff, K. Farias, J. P. Menzen, and G. Pessin, "Technological support for detection and prediction of plant diseases: A systematic mapping study," *Comput. Electron. Agric.*, vol. 181, p. 105922, Feb. 2021, doi: 10.1016/j.compag.2020.105922.

[17] E. Li, L. Wang, Q. Xie, R. Gao, Z. Su, and Y. Li, "A novel deep learning method for maize disease identification based on small sample-size and complex background datasets," *Ecol. Inform.*, vol. 75, p. 102011, Jul. 2023, doi: 10.1016/j.ecoinf.2023.102011.

[18] W. Li *et al.*, "Grape Disease Detection Using Transformer-Based Integration of Vision and Environmental Sensing," *Agronomy*, vol. 15, no. 4, p. 831, Apr. 2025, doi: 10.3390/agronomy15040831.

[19] J. Poyatos, D. Molina, A. D. Martinez, J. Del Ser, and F. Herrera, "EvoPruneDeepTL: An evolutionary pruning model for transfer learning based deep neural networks," *Neural Netw.*, vol. 158, pp. 59–82, Jan. 2023, doi: 10.1016/j.neunet.2022.10.011.

[20] "A Novel Approach to Detect Plant Disease Using DenseNet-121 Neural Network | springerprofessional.de." Accessed: Jan. 22, 2026. [Online]. Available: https://link.springer.com/chapter/10.1007/978-981-16-9967-2_7

[21] B. Liu, Y. Zhang, D. He, and Y. Li, "Identification of Apple Leaf Diseases Based on Deep Convolutional Neural Networks," *Symmetry*, vol. 10, no. 1, p. 11, Jan. 2018, doi: 10.3390/sym10010011.

[22] I. Ahmad, M. Hamid, S. Yousaf, S. T. Shah, and M. O. Ahmad, "Optimizing Pretrained Convolutional Neural Networks for Tomato Leaf Disease Detection," *Complexity*, vol. 2020, pp. 1–6, Sep. 2020, doi: 10.1155/2020/8812019.

[23] W. Zeng and M. Li, "Crop leaf disease recognition based on Self-Attention convolutional neural network," *Comput. Electron. Agric.*, vol. 172, p. 105341, May 2020, doi: 10.1016/j.compag.2020.105341.

[24] A. Jafar, N. Bibi, R. A. Naqvi, A. Sadeghi-Niaraki, and D. Jeong, "Revolutionizing agriculture with artificial intelligence: plant disease detection methods, applications, and their limitations," *Front. Plant Sci.*, vol. 15, Mar. 2024, doi: 10.3389/fpls.2024.1356260.

[25] M. A. Jasim and J. M. AL-Tuwaijari, "Plant Leaf Diseases Detection and Classification Using Image Processing and Deep Learning Techniques," *2020 Int. Conf. Comput. Sci. Softw. Eng. CSASE*, pp. 259–265, Apr. 2020, doi: 10.1109/CSASE48920.2020.9142097.

[26] A. Guerrero-Ibañez and A. Reyes-Muñoz, "Monitoring Tomato Leaf Disease through Convolutional Neural Networks," *Electronics*, vol. 12, no. 1, p. 229, Jan. 2023, doi: 10.3390/electronics12010229.
[27] I. Ahmed and P. K. Yadav, "A systematic analysis of machine learning and deep learning based approaches for identifying and diagnosing plant diseases," *Sustain. Oper. Comput.*, vol. 4, pp. 96–104, Jan. 2023, doi: 10.1016/j.susoc.2023.03.001.
[28] "Deep Transfer Learning Technique for Multimodal Disease Classification in Plant Images - Balaji - 2023 - Contrast Media & Molecular Imaging - Wiley Online Library." Accessed: Jan. 22, 2026. [Online]. Available: https://onlinelibrary.wiley.com/doi/10.1155/2023/5644727
[29] M. Kirola, K. Joshi, S. Chaudhary, N. Singh, H. Anandaram, and A. Gupta, "Plants Diseases Prediction Framework: A Image-Based System Using Deep Learning," in *2022 IEEE World Conference on Applied Intelligence and Computing (AIC)*, Jun. 2022, pp. 307–313. doi: 10.1109/AIC55036.2022.9848899.
[30] S. C. Gopi and H. Kishan Kondaveeti, "Transfer Learning for Rice Leaf Disease Detection," in *2023 Third International Conference on Artificial Intelligence and Smart Energy (ICAIS)*, Feb. 2023, pp. 509–515. doi: 10.1109/ICAIS56108.2023.10073711.
[31] Y. M. Abd Algani, O. J. Marquez Caro, L. M. Robladillo Bravo, C. Kaur, M. S. Al Ansari, and B. Kiran Bala, "Leaf disease identification and classification using optimized deep learning," *Meas. Sens.*, vol. 25, p. 100643, Feb. 2023, doi: 10.1016/j.measen.2022.100643.
[32] G. Dai, J. Fan, Z. Tian, and C. Wang, "PPLC-Net:Neural network-based plant disease identification model supported by weather data augmentation and multi-level attention mechanism," *J. King Saud Univ. - Comput. Inf. Sci.*, vol. 35, no. 5, p. 101555, May 2023, doi: 10.1016/j.jksuci.2023.101555.
[33] "Tomato Leaf Disease Detection and Classification Using Cnn | Mathematical Statistician and Engineering Applications." Accessed: Jan. 22, 2026. [Online]. Available: https://www.philstat.org/index.php/MSEA/article/view/853
[34] M. G. Yigezu, M. M. Woldeyohannis, and A. L. Tonja, "Early Ginger Disease Detection Using Deep Learning Approach," M. L. Berihun, Ed., in Lecture Notes of the Institute for Computer Sciences, Social Informatics and Telecommunications Engineering, vol. 411. Cham: Springer International Publishing, 2022, pp. 480–488. doi: 10.1007/978-3-030-93709-6_32.
[35] D. Tirkey, K. K. Singh, and S. Tripathi, "Performance analysis of AI-based solutions for crop disease identification, detection, and classification," *Smart Agric. Technol.*, vol. 5, p. 100238, Oct. 2023, doi: 10.1016/j.atech.2023.100238.
[36] İ. Yağ and A. Altan, "Artificial Intelligence-Based Robust Hybrid Algorithm Design and Implementation for Real-Time Detection of Plant Diseases in Agricultural Environments," *Biology*, vol. 11, no. 12, p. 1732, Dec. 2022, doi: 10.3390/biology11121732.
[37] R. Sujatha, J. M. Chatterjee, N. Jhanjhi, and S. N. Brohi, "Performance of deep learning vs machine learning in plant leaf disease detection," *Microprocess. Microsyst.*, vol. 80, p. 103615, Feb. 2021, doi: 10.1016/j.micpro.2020.103615.
[38] E. C. Too, L. Yujian, S. Njuki, and L. Yingchun, "A comparative study of fine-tuning deep learning models for plant disease identification," *Comput. Electron. Agric.*, vol. 161, pp. 272–279, Jun. 2019, doi: 10.1016/j.compag.2018.03.032.
[39] J. A. Pandian, V. D. Kumar, O. Geman, M. Hnatiuc, M. Arif, and K. Kanchanadevi, "Plant Disease Detection Using Deep Convolutional Neural Network," *Appl. Sci.*, vol. 12, no. 14, p. 6982, Jan. 2022, doi: 10.3390/app12146982.
[40] J. Chen, J. Chen, D. Zhang, Y. Sun, and Y. A. Nanehkaran, "Using deep transfer learning for image-based plant disease identification," *Comput. Electron. Agric.*, vol. 173, p. 105393, Jun. 2020, doi: 10.1016/j.compag.2020.105393.
[41] P. Sharma, Y. P. S. Berwal, and W. Ghai, "Performance analysis of deep learning CNN models for disease detection in plants using image segmentation," *Inf. Process. Agric.*, vol. 7, no. 4, pp. 566–574, Dec. 2020, doi: 10.1016/j.inpa.2019.11.001.
[42] B. Tugrul, E. Elfatimi, and R. Eryigit, "Convolutional Neural Networks in Detection of Plant Leaf Diseases: A Review," *Agriculture*, vol. 12, no. 8, p. 1192, Aug. 2022, doi: 10.3390/agriculture12081192.
[43] S. M. Hassan, A. K. Maji, M. Jasiński, Z. Leonowicz, and E. Jasińska, "Identification of Plant-Leaf Diseases Using CNN and Transfer-Learning Approach," *Electronics*, vol. 10, no. 12, p. 1388, Jan. 2021, doi: 10.3390/electronics10121388.
[44] K. P. Ferentinos, "Deep learning models for plant disease detection and diagnosis," *Comput. Electron. Agric.*, vol. 145, pp. 311–318, Feb. 2018, doi: 10.1016/j.compag.2018.01.009.
[45] W. Shafik, A. Tufail, C. De Silva Liyanage, and R. A. A. H. M. Apong, "Using transfer learning-based plant disease classification and detection for sustainable agriculture," *BMC Plant Biol.*, vol. 24, no. 1, p. 136, Feb. 2024, doi: 10.1186/s12870-024-04825-y.
[46] A. Khakimov, I. Salakhutdinov, A. Omolikov, and S. Utaganov, "Traditional and current-prospective methods of agricultural plant diseases detection: A review," *IOP Conf. Ser. Earth Environ. Sci.*, vol. 951, no. 1, p. 012002, Jan. 2022, doi: 10.1088/1755-1315/951/1/012002.
[47] "New Plant Diseases Dataset." Accessed: Jan. 13, 2026. [Online]. Available: https://www.kaggle.com/datasets/vipoooool/new-plant-diseases-dataset
[48] S. S. Harakannanavar, J. M. Rudagi, V. I. Puranikmath, A. Siddiqua, and R. Pramodhini, "Plant leaf disease detection using computer vision and machine learning algorithms," *Glob. Transit. Proc.*, vol. 3, no. 1, pp. 305–310, Jun. 2022, doi: 10.1016/j.gltp.2022.03.016.
[49] "shkelqimsherifi/AI_DeepLearning_CNN_Model_Plant_Disease_Detaction," GitHub. Accessed: Feb. 17, 2026. [Online]. Available: https://github.com/shkelqimsherifi/AI_DeepLearning_CNN_Model_Plant_Disease_Detaction
[50] M. D. Zeiler and R. Fergus, "Visualizing and Understanding Convolutional Networks," in *Computer Vision – ECCV 2014*, D. Fleet, T. Pajdla, B. Schiele, and T. Tuytelaars, Eds., Cham: Springer International Publishing, 2014, pp. 818–833. doi: 10.1007/978-3-319-10590-1_53.
[51] L. T. Ramos and A. D. Sappa, "A Decade of You Only Look Once (YOLO) for Object Detection: A Review," Aug. 03, 2025, *arXiv*: arXiv:2504.18586. doi: 10.48550/arXiv.2504.18586.
[52] "Models and pre-trained weights — Torchvision main documentation." Accessed: Feb. 07, 2026. [Online]. Available: https://docs.pytorch.org/vision/main/models.html
[53] S. P. Mohanty, D. P. Hughes, and M. Salathé, "Using Deep Learning for Image-Based Plant Disease Detection," *Front. Plant Sci.*, vol. 7, p. 1419, Sep. 2016, doi: 10.3389/fpls.2016.01419.
[54] S. Cheng *et al.*, "A High Performance Wheat Disease Detection Based on Position Information," *Plants*, vol. 12, no. 5, Mar. 2023, doi: 10.3390/plants12051191.
[55] A. Bhargava, A. Shukla, O. P. Goswami, M. H. Alsharif, P. Uthansakul, and M. Uthansakul, "Plant Leaf Disease Detection, Classification, and Diagnosis Using Computer Vision and Artificial Intelligence: A Review," *IEEE Access*, vol. 12, pp. 37443–37469, 2024, doi: 10.1109/ACCESS.2024.3373001.
[56] W. B. Demilie, "Plant disease detection and classification techniques: a comparative study of the performances," *J. Big Data*, vol. 11, no. 1, p. 5, Jan. 2024, doi: 10.1186/s40537-023-00863-9.
[57] M. Arsenovic, M. Karanovic, S. Sladojevic, A. Anderla, and D. Stefanovic, "Solving Current Limitations of Deep Learning Based Approaches for Plant Disease Detection," *Symmetry*, vol. 11, no. 7, p. 939, Jul. 2019, doi: 10.3390/sym11070939.
[58] K. Simonyan and A. Zisserman, "Very Deep Convolutional Networks for Large-Scale Image Recognition," Apr. 10, 2015, *arXiv*: arXiv:1409.1556. doi: 10.48550/arXiv.1409.1556.
[59] C. Szegedy, V. Vanhoucke, S. Ioffe, J. Shlens, and Z. Wojna, "Rethinking the Inception Architecture for Computer Vision," in *2016 IEEE Conference on Computer Vision and Pattern Recognition (CVPR)*, Jun. 2016, pp. 2818–2826. doi: 10.1109/CVPR.2016.308.
[60] M. Ahmad, M. Abdullah, H. Moon, and D. Han, "Plant Disease Detection in Imbalanced Datasets Using Efficient Convolutional Neural Networks With Stepwise Transfer Learning," *IEEE Access*, vol. 9, pp. 140565–140580, 2021, doi: 10.1109/ACCESS.2021.3119655.
[61] "Plant Disease Classification and Adversarial Attack Using SimAM-EfficientNet and GP-MI-FGSM." Accessed: Jan. 22, 2026. [Online]. Available: https://www.mdpi.com/2071-1050/15/2/1233
[62] A. G. Howard *et al.*, "MobileNets: Efficient Convolutional Neural Networks for Mobile Vision Applications," Apr. 17, 2017, *arXiv*: arXiv:1704.04861. doi: 10.48550/arXiv.1704.04861.
[63] A. Krizhevsky, I. Sutskever, and G. E. Hinton, "ImageNet Classification with Deep Convolutional Neural Networks," in *Advances in Neural Information Processing Systems*, Curran Associates, Inc., 2012. Accessed: Jan. 22, 2026. [Online]. Available: https://proceedings.neurips.cc/paper_files/paper/2012/hash/c399862d3b9d6b76c8436e924a68c45b-Abstract.html

[64] “Early Detection of Plant Viral Disease Using Hyperspectral Imaging and Deep Learning.” Accessed: Jan. 22, 2026. [Online]. Available: https://www.mdpi.com/1424-8220/21/3/742
[65] K. He, X. Zhang, S. Ren, and J. Sun, “Deep Residual Learning for Image Recognition,” in *2016 IEEE Conference on Computer Vision and Pattern Recognition (CVPR)*, Las Vegas, NV, USA: IEEE, Jun. 2016, pp. 770–778. doi: 10.1109/CVPR.2016.90.
[66] M. Ouhami, A. Hafiane, Y. Es-Saady, M. El Hajji, and R. Canals, “Computer Vision, IoT and Data Fusion for Crop Disease Detection Using Machine Learning: A Survey and Ongoing Research,” *Remote Sens.*, vol. 13, no. 13, p. 2486, Jan. 2021, doi: 10.3390/rs13132486.